\documentclass[10pt,twocolumn,twoside]{IEEEtran}

\usepackage{booktabs} 
\usepackage{amsfonts}
\usepackage{amsmath}
\usepackage{amssymb}
\usepackage{bm}
\usepackage{url}
\usepackage{graphicx}
\usepackage{subcaption}
\usepackage{multirow}
\usepackage{float}
\usepackage{balance} 
\usepackage{array}
\usepackage{graphicx}
\usepackage{caption}
\usepackage{comment}
\usepackage{booktabs}
\usepackage{multirow}
\newcommand*\rot{\rotatebox{90}}
\newcolumntype{L}[1]{>{\raggedright\let\newline\\\arraybackslash\hspace{0pt}}m{#1}}

\newcommand{\eat}[1]{} 
\begin{document}

\title{Multi-layer Feature Aggregation for Deep Scene Parsing Models}
\author{Litao~Yu, 
	Yongsheng~Gao,~\IEEEmembership{Senior~Member,~IEEE,}
	Jun~Zhou,~\IEEEmembership{Senior~Member,~IEEE,}
	Jian~Zhang,~\IEEEmembership{Senior~Member,~IEEE,}
	Qiang~Wu,~\IEEEmembership{Senior~Member,~IEEE}
\thanks{Litao Yu (litao.yu@uts.edu.au), Jian Zhang (jian.zhang@uts.edu.au) and Qiang Wu (qiang.wu@uts.edu.au) are with Global Big Data Technologies Centre, University of Technology Sydney, Ultimo, 2007, NSW, Australia. Yongsheng Gao (yongsheng.gao@griffith.edu.au) and Jun Zhou (jun.zhou@griffith.edu.au) are with the Institute for Integrated and Intelligent Systems, Griffith University, Nathan, 4111, QLD, Australia.  This work was supported by the Australian Research Council under Discovery Grants DP140101075 and DP180100958.}
 \thanks{Manuscript received July 1, 2020.}
}
 
\maketitle
 
\markboth{IEEE Transactions on Neural Networks and Learning Systems}%
 {Yu \MakeLowercase{\textit{et al.}}: Multi-layer Feature Aggregation for Deep Scene Parsing Models}

\begin{abstract}
Scene parsing from images is a fundamental yet challenging problem in visual content understanding. In this dense prediction task, the parsing model assigns every pixel to a categorical label, which requires the contextual information of adjacent image patches. So the challenge for this learning task is to simultaneously describe the geometric and semantic properties of objects or a scene. In this paper, we explore the effective use of multi-layer feature outputs of the deep parsing networks for spatial-semantic consistency by designing a novel feature aggregation module to generate the appropriate global representation prior, to improve the discriminative power of features. The proposed module can auto-select the intermediate visual features to correlate the spatial and semantic information. At the same time, the multiple skip connections form a strong supervision, making the deep parsing network easy to train. Extensive experiments on four public scene parsing datasets prove that the deep parsing network equipped with the proposed feature aggregation module can achieve very promising results.
\end{abstract}


\section{Introduction}

Scene parsing based on semantic segmentation is a dense classification task for visual content analysis in image processing. The goal is to assign a class label to every single pixel in given images, i.e., parse a scene into different geometric regions associated with semantic categories such as \emph{sky, road} and \emph{bicycles}. This topic has drawn a broad research interest for many applications such as surveillance for security \cite{STAP:STIS}, robot sensing \cite{ICRA14:ROBOT_SENSING} and auto-navigation \cite{CVPR17:NAV}.

The difficulty of unconstrained semantic segmentation mainly lies in the high varieties of scenes and their associated labels. Some categories are semantically confusing due to spatial-semantic inconsistencies. For example, regions of ``pedestrians'' and ``riders'' are often indistinguishable, and ``cars'' are usually affected by visual scales, occlusions and illuminations. Therefore, the spatial and semantic information shall be consistent to address this challenge. Furthermore, accurate label prediction at the pixel level requires high resolution of visual feature representations. For example, in the challenging Cityscapes dataset \cite{CVPR16:CITYSCAPES}, it is comparably easy to segment some large objects such as ``road'' and ``building'', but very difficult to localize and sketch the contours of small objects such as ``poles'' and ``traffic signs''. 

Recently, the development of deep convolutional neural networks has led to remarkable progress in scene parsing due to their powerful feature representation ability to describe the local visual properties. Deep parsing networks are often fine-tuned based on the pre-trained classification networks, e.g., deep residual networks \cite{CVPR16:RESNET}. These classification networks usually stack convolution and down-sampling layers to obtain visual feature maps with rich semantics. The deeper layer features with rich semantics are crucial for accurate classification, but lead to the reduced resolution and in turn spatial information loss. Such information loss is detrimental for scene parsing because it decreases the localization accuracy. On the other hand, the spatial-sensitive feature maps in the shallow layers are rarely optimized due to the vanishing gradient. Thus, how to keep the spatial-semantic consistencies becomes an open problem in scene parsing. To address this issue, several modifications to Fully Convolutional Networks (FCNs) \cite{CVPR15:FCN} have been made. For example, in the encoder-decoder structure such as UNet \cite{MICCAI15:UNET}, the encoder maps the original images into low-resolution feature representations, while the decoder mainly restores the spatial information with skip-connections. Unfortunately, the missing geometric information cannot be fully restored. Another popular method that has been widely used in segmentation is the dilated (atrous) convolution \cite{ARXIV:DILATED}, which can enlarge the receptive field in the feature maps without adding more computation overhead, thus more visual details are preserved. Combining the encoder-decoder structure and dilated convolution can effectively boost the pixel-wise prediction accuracy \cite{TPAMI:DEEPLAB}, but is extremely computational demanding.

\begin{figure*}[t]
\centering
    \begin{minipage}{0.4\textwidth}
        \includegraphics[width=1\textwidth]{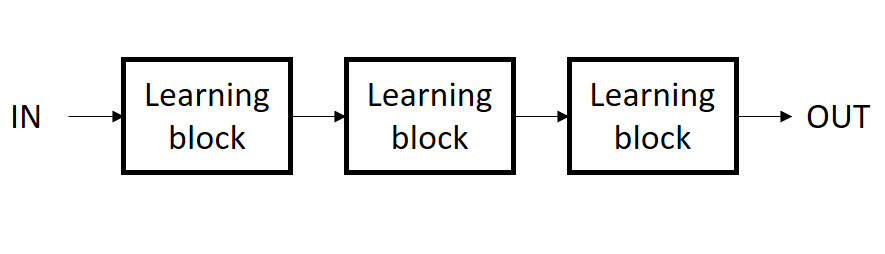}
        \subcaption{Sequential (no aggregation)}
	   \label{FIG:NA}
    \end{minipage}
    ~
    \begin{minipage}{0.4\textwidth}
        \includegraphics[width=1\textwidth]{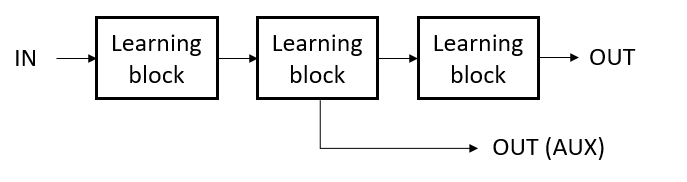}
        \subcaption{Auxiliary loss}
	   \label{FIG:AL}
    \end{minipage}
   ~
    \begin{minipage}{0.4\textwidth}
        \includegraphics[width=1\textwidth]{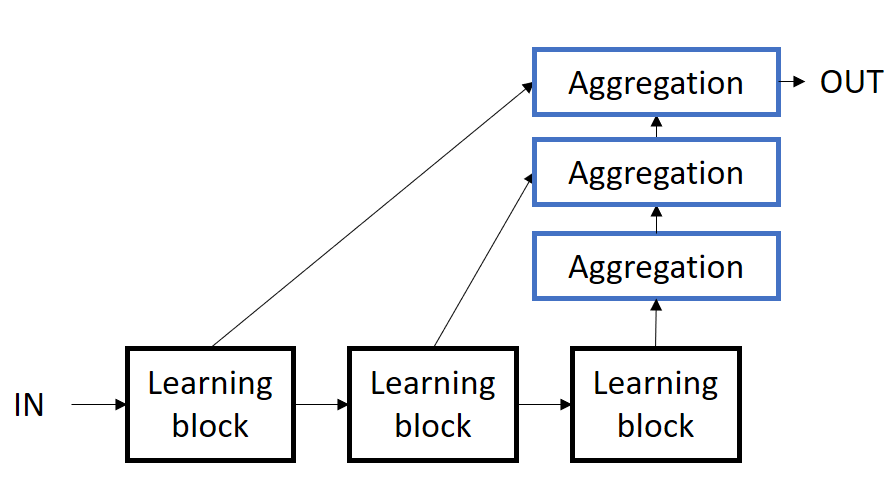}
        \subcaption{Skip-connection}
	   \label{FIG:CA}
    \end{minipage}
    ~
    \begin{minipage}{0.4\textwidth}
        \includegraphics[width=1\textwidth]{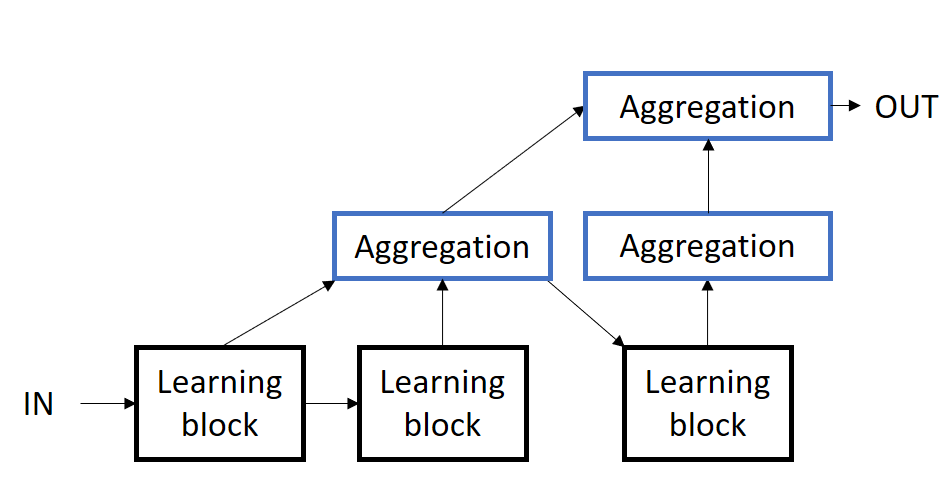}
        \subcaption{Deep layer aggregation}
	   \label{FIG:DA}
    \end{minipage}    
\caption{Different approaches to utilize multiple-layer outputs. (a) Sequential learning blocks without aggregation as default for supervised learning tasks. (b) Sequential learning blocks with an auxiliary loss branch. (c) Skip-connections through multiple learning blocks \cite{CVPR17:DENSENET}. (d) Deep layer aggregation proposed in \cite{CVPR18:DLA}. The ``learning block'' refers to simple or composite convolutional structures (e.g., residual block \cite{CVPR16:RESNET}). The ``aggregation'' here means either the residual-add or simple concatenation on the channel axis.}
\label{FIG:4AGGREGATIONS}
\end{figure*}

In deep parsing networks, the shallow convolution features, i.e., the early convolution outputs, encode the low-level spatial visual information such as edges, corners and circles. The high-level features in the deep blocks, on the other hand, carry more semantic information, including instance and category-level evidences, but lack geometric information. The outputs in the middle convolution blocks carry miscellaneous spatial and semantic properties of images, forming the mid-level feature maps. In the optimization procedure, the gradients can be easily passed through the deeper convolution blocks, but the weights in the shallower convolution blocks are rarely updated, making a very slow convergence. At the same time, the different layer outputs of feature representations are complementary to each other, which should be re-considered for any dense prediction tasks. However, the effective aggregation of multiple feature outputs is rarely recognized as a critical step in model optimization.

In this paper, we propose a novel feature aggregation module applied on multiple layer outputs of deep parsing networks to effectively capture the long-range contextual information. In the proposed method, the module can auto-select the most useful feature maps to form a discriminative visual feature representation for dense prediction. Thus, with the neatly designed feature aggregation module, the proposed parsing network, named {\bf S}patial-semantic {\bf A}ggregation {\bf Net}work (SANet), can take the long-contextual information of deep parsing networks into consideration, which has the following two advantages: (1) it uses multiple feature maps to form a strong supervision, in which the spatial-semantic properties are automatically correlated; and (2) the gradients are directly passed into multiple layers, making the parsing network easy to train. In the experiment, the proposed SANet achieves very promising performance on four widely used benchmark datasets, i.e., NYU Depth v2 \cite{ECCV12:NYU}, SUN RGB-D \cite{NIPS14:SUN}, ADE20K \cite{CVPR17:ADE20K} and Cityscapes \cite{CVPR16:CITYSCAPES}. The spatial-semantic feature aggregation module with the long-contextual information paradigm can be extended to other image processing models to benefit the spatial-aware learning tasks.

The rest of the paper is organized as follows. Section \ref{SEC:RELATED_WORK} introduces related work. Section \ref{SEC:METHOD} elaborates the proposed feature aggregation module as well as the learning framework of SANet for scene parsing. Experimental results and analysis are presented in Section \ref{SEC:EXP}. Finally, Section \ref{SEC:CONCLUSION} concludes the paper.

\section{Related work}
\label{SEC:RELATED_WORK}

\subsection{Deep learning based parsing networks}

Deep parsing models based on fully-convolutional networks \cite{CVPR15:FCN} have achieved significant outcomes on large-scale benchmark datasets \cite{CVPR16:CITYSCAPES,CVPR17:ADE20K}. Following the first deep parsing model FCN \cite{CVPR15:FCN}, DeconvNet \cite{CVPR15:DECONV}, SegNet \cite{TPAMI:SEGNET}, UNet \cite{MICCAI15:UNET} and their variants \cite{ARXIV:CONVCRF,CVPR17:GFRN,ARXIV:DECODER} adopt the encoder-decoder framework, where skip-connections are frequently used to refine segmentation masks. The aim to use skip-connections is to recover the geometric information that is represented in the early convolution blocks, but such aggregations fail to effectively select the most appropriate feature components because they are mixed with higher semantic information. Moreover, it is not clear that how much information can help to preserve the geometric information and aggregate the semantics in the 2D feature maps. For scene parsing tasks, exploring the contextual information is beneficial for semantic understanding, which requires large receptive fields with adjacent pattern information in deep models. Based on this consideration, some methods use dilated convolutions to replace traditional convolution layers to enlarge the receptive field \cite{ARXIV:DILATED}, or some others adopt graphical models with effective inference to analyze the surrounding visual information \cite{TPAMI:DAG}. The multi-scale processing technique is commonly used in many learning tasks such as visual localization and detection. Specific to semantic segmentation, employing multi-scale inputs \cite{CVPR17:REFINENET,CVPR16:ATTENTION_SCALE} or applying multi-scale aggregation \cite{CVPR17:PSPNET} are effective ways to improve the model accuracy. In \cite{CVPR19:AUTODEEPLAB}, the authors proposed to use neural architecture search to discover the deep model structures for semantic segmentation, which is no longer dependent on the pre-trained deep models and can obtain satisfactory results. However, its generalization capability remains an issue because the architecture is searched on a specific validation dataset rather than a universal one, so it is hard to adapt to new data environments.

\subsection{Effective utilization of multiple layer outputs}

Deep neural networks have been extensively used for various image processing tasks. Specifically, the pre-trained deep networks on large-scale datasets serve as stem networks for new learning tasks. The stem networks usually contain multiple learning blocks such as residual blocks in deep residual networks \cite{CVPR16:RESNET}. The densely connected network \cite{CVPR17:DENSENET} is a canonical architecture for semantic fusion, which is designed to better propagate features and losses through skip-connections that concatenate all the feature maps in a learning block. For spatial-aware learning tasks such as object detection, feature fusion \cite{CVPR17:FPN} is designed to equalize and standardize semantics across the levels of a pyramid feature hierarchy through top-down and lateral connections. 

With the powerful GPUs, the number of layers in the stem networks can be easily extended to several hundreds or even more than one thousand, but the resulting performance does not increase linearly. In a deep convolution network, the stacked learning blocks are divided into stages according to the resolution and representation capacity of feature maps. The feature map outputs in deeper stages contain more semantic information but are spatially coarser than shallow layers. Hence, further exploration is needed on how to connect these layers or learning blocks. In Figure \ref{FIG:4AGGREGATIONS}, we illustrate four feature inference schemes. The basic architecture (a) without layer aggregation or branching structure is widely used in supervised learning tasks \cite{CVPR16:RESNET,CVPR18:SENET, CVPR15:FCN, ARXIV:DILATED,NIPS15:FASTER_RCNN,CVPR15:CAP}. To address the spatial-aware learning task, deep semantic segmentation models\cite{MICCAI15:UNET,TPAMI:DEEPLAB} adopt the skip-connections (b) across multiple learning stages, fusing the feature maps to restore the spatial information. The sequential inference with auxiliary loss (c) was firstly used in \cite{CVPR15:GOOGLENET} for very deep neural networks to enhance the supervision, and was also used in PSPNet \cite{CVPR17:PSPNET} for semantic segmentation. In \cite{CVPR18:DLA}, the authors designed a multi-stage aggregation without losing any intermediate feature representations to improve the model performance, as is illustrated in (d). However, the design of the aggregation relies on experience, and the basic sequential structure of the stem network has to be modified, which means the model needs to be trained from the very beginning. In our proposed aggregation structure, we design a spatial-semantic aggregation module to effectively aggregate the feature maps without changing the stem network structure.

\section{Method}
\label{SEC:METHOD}

In this section, we start with the observation and analysis of skip-connections and auxiliary losses when applying FCN for scene parsing, which motivate the design of the multi-layer feature aggregation module, then we give the details of the whole learning framework.

\subsection{Multi-layer feature aggregation}
\label{SUBSEC:MLFA}

To effectively correlate the spatial and semantic information, the feature aggregation should reserve the discriminative feature components and abandon the useless ones. Specific to a FCN, the early convolution outputs have dual functionalities: describing the geometric properties and acquiring new knowledge in deeper stages. The features of the middle and late layers are less spatial-aware but semantic indicative. To effectively use multiple feature maps from different layers in a FCN, skip-connections and auxiliary losses are commonly used. Skip-connections are usually used in deep learning-based scene parsing models such as DeepLab \cite{TPAMI:DEEPLAB} to correlate the spatial information in the early convolution layers and the semantics in the late feature outputs. Auxiliary losses, on the other hand, can help improve the discriminative power by adding extra output branches in intermediate layers and providing stronger supervision.

To observe the effectiveness of skip-connections and auxiliary losses in FCN, we conducted a simple indoor scene parsing experiment on two small datasets NYU Depth v2 \cite{ECCV12:NYU} (40 classes) and SUN RGB-D \cite{NIPS14:SUN} (37 classes). We used the ResNeXt50 model \cite{CVPR17:RESNEXT} as the stem network, replacing the global average pooling layer and classification layer with two convolution layers to form a pixel-wise classifier for dense prediction.

We selected the following layer outputs to skip-connect the last feature map before the final convolution (pixel classifier): (a) the ReLU layer after the first convolution ({\bf s0}); (b) the final ReLU layers at each of the four stages before resolution changes ({\bf s1}, {\bf s2}, {\bf s3} and {\bf s4}). Similarly, we added dense layers on these layer outputs as auxiliary losses to observe their effectiveness. In the training procedure, we used the IoU (intersection over union) to evaluate the performance on the validation (test) set. The results of different skip-connections and auxiliary losses are summarised in Table \ref{TB:SKIP}.

\begin{table}[h]
\centering 
\caption{Comparison of IoU of FCNs with different skip-connections ({\bf SC}) and auxiliary losses ({\bf AL}) on NYU Depth v2 and SUN RGB-D validation sets.}
\label{TB:SKIP}
\begin{tabular}{|c||c|c|c|c|}
\hline
\multirow{2}{*}{{\bf Layer}} & \multicolumn{2}{c|}{NYU Depth v2} & \multicolumn{2}{c|}{SUN RGB-D}	 \\
\cline{2-5}  & SC &AL  & SC &AL  \\
\hline

{\bf None}     & \multicolumn{2}{c|}{40.2} & \multicolumn{2}{c|}{40.6} \\
\cline{2-5}
{\bf s0}	     & 39.2 & 40.4   & 39.8 & 40.1 \\
{\bf s1}	      & 40.0 & 40.4  & {\bf 40.4} & 40.5 \\
{\bf s2}	      & 39.6 & {\bf 40.6}   & 40.3 & {\bf 40.7} \\
{\bf s3}	       & 39.4 & 40.4 & 40.2  & 40.6 \\
{\bf s4}	       & {\bf 40.4} & 40.2 & 40.3  & 40.6 \\
\hline
\end{tabular}
\end{table}

Inspecting the validation results in the table above, we can observe that some skip-connections or auxiliary losses can improve the classification accuracy, while some others cannot. Furthermore, applying the two approaches to the early convolution output ({\bf s0} in the experiment) essentially decreases the IoU. For a given stem network with very deep structures and a large-scale dataset, it is hard to select the proper intermediate layers to form the most appropriate feature representation. Based on such observations, we design a multi-layer feature selection method to fully utilize multiple feature maps thus improve the scene parsing performance.  

Instead of concatenating multiple-layer outputs as is in skip-connections or applying an auxiliary loss in an intermediate layer output, we design a nonlinear feature aggregation method throughout the whole stem network, without wasting any layer information, to better fit the potential data distribution.

We consider the multi-layer outputs in a FCN as a sequence of feature maps. Given the labels as supervision information, the global feature representation should well leverage the spatial and semantic properties for the spatial-aware prediction. Therefore, the layer dependency in the sequence of feature maps should be modelled to learn the spatial-aware feature representation. Here we use the long short-term memory (LSTM) as a feature selection function conducted on multi-layer outputs. LSTM can model the long-term dependencies in a sequence. Consequently, it acts as a feature selection function to form the appropriate feature representation. The core part of LSTM is a memory cell $\mathbf{c}_t$ at the time step $t$ that records the history of input sequence observed up to that time step, and the behaviour of the cell is controlled by an input gate $\mathbf{i}_t$, a forget gate $\mathbf{f}_t$ and an output gate $\mathbf{o}_t$. In a fully-connected LSTM (FC-LSTM), these three gates are computed by affine mappings with non-linear activations, which control whether to forget the current cell value if it should read its input and whether to output a new cell value.

The major limitation of FC-LSTM in modelling sequential data is the usage of full connections in input-to-state and state-to-state transitions in which no spatial information is encoded. Moreover, the affine mappings are with full weight matrices, leading to high computational complexity and the over-fitting problem. To address this issue, we use 2D convolution LSTM \cite{NIPS15:CONVLSTM} (ConvLSTM) instead of FC-LSTM to aggregate the multi-layer feature outputs. 

Suppose the 2D feature outputs of multiple learning blocks are $\mathbf{x}_1, \ldots, \mathbf{x}_m$, where $m$ is the number of feature map candidates. The $t$-th feature map ($1\le t \le m$) is a 3D tensor, i.e., $\mathbf{x}_t\in\mathbb{R}^{w_t\times h_t\times c_t}$, where $w_t$, $h_t$ and $c_t$ are the width, height and number of channels, respectively. To feed the 2D feature outputs into the ConvLSTM, all feature maps should be converted to the identical-shaped tensors $\mathbf{x}'_1, \ldots, \mathbf{x}'_m$ having the same resolution and dimension. Assume shape of the target feature map is $w'\times h'\times c'$, the change of the resolution from $w_t\times h_t$ to $w'\times h'$ is implemented by down-sampling, and the channel conversion from $c_t$ to $c'$ is computed by $1\times 1$ convolution with no bias. When $c_t>c'$, the $1\times 1$ convolution acts as the dimensionality reduction, otherwise it expands the original features into a higher feature space to better fit the high-dimenional non-linearity. Formally, the mapping from $\mathbf{x}_t \in \mathbb{R}^{w_t\times h_t\times c_t}$ to $\mathbf{x}'_t \in \mathbb{R}^{w'\times h'\times c'}$ is as follows:
\begin{equation}
\mathbf{x}'_t = \mathbf{w}_t * \psi(\mathbf{x}_t)
\end{equation}
where $*$ denotes the 2D convolution, $\mathbf{w}_t$ is the convolution kernel, and $\psi(\cdot)$ is the down-sampling operator\footnote{We use the bilinear interpolation on the 2D grids to adjust the feature resolution.}, respectively.

\begin{figure*}[t!]
\centering
    \includegraphics[width=0.8\textwidth]{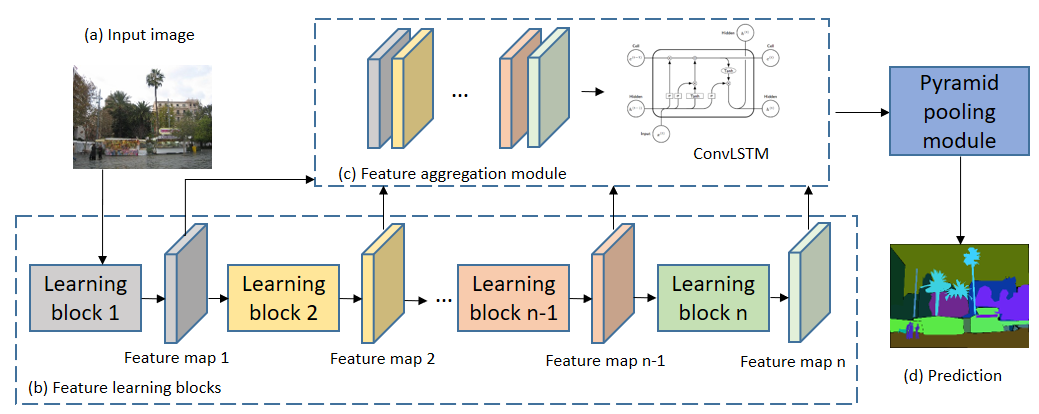}   
    \caption{Overview of the deep parsing framework with a feature aggregation module. Given an input image (a), we use a pre-trained deep model (b) as a stem network for feature inference. At the same time, we keep the output feature maps of multiple learning blocks and send them into a feature aggregation module (c) for spatial-semantic consistencies. Finally, the aggregated feature representation goes through a pyramid pooling module to get the final per-pixel prediction (d).}
\label{FIG:FRM}
\end{figure*}

The feature map $\mathbf{x}'_t$ can be considered as $w_t\times h_t$ feature vectors standing on a spatial grid. The ConvLSTM learns the sequential state of a certain cell in the grid by the inputs and past states of its local neighbours, which is implemented by using a convolution operator in the state-to-state and input-to-state transitions. Similar to FC-LSTM, the input gate $\mathbf{i}_t$ controls whether ConvLSTM considers the current input $\mathbf{x}'_t$, the forget gate $\mathbf{f}_t$ controls whether ConvLSTM forgets the previous memory $\mathbf{c}_{t-1}$, and the output gate $\mathbf{o}_t$ controls how much information will be read from memory $\mathbf{c}_t$ to the current hidden state $\mathbf{h}_t$. The computation stream of ConvLSTM is as follows:

\begin{align}
\mathbf{i}_t & = \sigma(\mathbf{w}_{ix}*\mathbf{x}'_t+ \mathbf{w}_{ih}*\mathbf{h}_{t-1}+\mathbf{b}_i),  \\
\mathbf{f}_t & = \sigma(\mathbf{w}_{fx}*\mathbf{x}'_t+\mathbf{w}_{fh}*\mathbf{h}_{t-1}+\mathbf{b}_f), \\
\mathbf{o}_t & = \sigma(\mathbf{w}_{ox}*\mathbf{x}'_t+\mathbf{w}_{oh}*\mathbf{h}_{t-1}+\mathbf{b}_o), \\
\mathbf{g}_t & = \tanh (\mathbf{w}_{gx}*\mathbf{x}'_t+\mathbf{w}_{gh} *\mathbf{h}_{t-1}+\mathbf{b}_g),  \\
\mathbf{c}_t & = \mathbf{f}_t \circ \mathbf{c}_{t-1} + \mathbf{i}_t\circ \mathbf{g}_t,  \\  
\mathbf{h}_t &= \mathbf{o}_t \circ \tanh (\mathbf{c}_t),
\end{align}
where $\sigma(\cdot)$ is the sigmoid function and $\circ$ is the element-wise multiplication, respectively. The convolution kernels $\mathbf{w}_{*x}$ and $\mathbf{w}_{*h}$ are the ConvLSTM state and recurrent transformations, and $\mathbf{b}_*$ are bias matrices.

Given a sequence of feature maps $\mathbf{x}'_1, \ldots, \mathbf{x}'_m$ from $m$ learning blocks, the output of 2D ConvLSTM is a sequence of tensors. The final feature map that correlates the spatial and semantic information is calculated as:

\begin{equation}
\mathbf{y}=\frac{1}{m}\sum\limits \text{ConvLSTM}(\mathbf{x}'_1, \ldots, \mathbf{x}'_m) . 
\end{equation}

Considering the specific case of pixel-level classification that requires larger receptive fields, a higher dilation rate (e.g., 2) is used in the ConvLSTM. By adding the ConvLSTM as a feature aggregation function on multiple-layer outputs, the deep parsing structure has the following three advantages. First, the feature aggregation module is able to keep the spatial-semantic consistencies, which is not a linear combination of multiple feature maps, but is an extra learning module to acquire new knowledge thus enhance the global feature representation. Second, the feature aggregation module uses multiple feature maps at different stages as the input through the stem network, forming a strong supervision and making it easier to train, because the gradients from the late layer for pixel-wise classification can be directly passed into shallower learning blocks. Thus, compared to the deep parsing networks such as \cite{CVPR17:PSPNET,CVPR19:DANET}, the convergence is faster when applying the feature aggregation module. Third, the feature aggregation module can be directly inserted in any existing FCN pipeline, which does not significantly increase the computational complexity overhead yet enhance the feature representation capability.

Recurrent modules have been widely used to improve the performance by considering the visual contextual information in some visual pattern recognition works \cite{CVPR18:RNN_SEG,CVPR19:TDBU,ECCV18:AAF,TPAMI:DAG}. For example, the ConvLSTM is used to help model the motion \cite{ISBI19:ML_CONVLSTM} and spatial-temporal dependencies in video frames \cite{BMVC18:FSS_CONVLSTM}. In our work, we use the recurrent module from another perspective for multi-layer feature aggregation for spatial-semantic consistencies, which differs from their contexts and it is the key contribution for spatial-aware learning tasks.  
In \cite{CVPR18:CCL}, the authors proposed to make multi-level context contrasted local features to aggregate multi-layer outputs. The key difference to our model is that they adopt the gated sum to control the information flow, while we use the recurrent model to auto-aggregate the features. The gated sum is computed multiple times for every two layer aggregation, so for the whole learning framework it is less computationally efficient compared to our method.

\subsection{The overall SANet learning architecture}

We now present the architecture of spatial-semantic aggregation network (SANet). The overall framework is illustrated in Figure \ref{FIG:FRM}. In the proposed framework, the global feature representation is composed of two parallel pipelines. The first pipeline (b) is essentially a dilated FCN that consists of multiple learning blocks, during which both the feature resolution and feature dimensionality are changed. The second pipeline (c) is in parallel with (b), in which multiple feature maps are aggregated by the spatial-semantic feature selection module. 

The stem network of SANet used in our work is a 50-layer ResNeXt \cite{CVPR17:RESNEXT}. Compared with ResNet proposed in \cite{CVPR16:RESNET}, ResNeXt adopts the aggregated transformation by increasing the cardinality, which can improve the classification accuracy without increasing the width of the bottleneck block or the depth of the whole network. The ResNeXt-50 has four learning stages (blocks) after the early convolution. At each stage the number of channels doubles while the resolution halves. We choose the final ReLU activation before resolution changes, so five feature map outputs are selected as the intermediate feature candidates to feed into the spatial-semantic feature aggregation module. Given an input image (a) in Figure \ref{FIG:FRM}, it has two paths to arrive at the final feature representation prior. The first path is the sequential learning blocks of ResNeXt-50 in (b), and the second path is the feature map conversion and a spatial-semantic feature aggregation module in (c), respectively. We then apply a pyramid feature module (PSP) \cite{CVPR17:PSPNET} as a global feature representation for the pixel-label prediction of the objectives. We do not apply auxiliary loss to any intermediate layer to enhance the discriminative power because the multi-layer feature aggregation module can already conduct the feature selection and pass the gradients into multiple layers in the back-forward optimization.

\begin{table*}[t!]
\centering 
\caption{Computaional complexity analysis.}
\label{TB:FLOPS}
\begin{tabular}{|c||c|c|c|c|c|}
\hline
{\bf Method}	& {\bf Stem network}	& {\bf Input size}	& {\bf Output} 	& {\bf FLOPs} & {\bf Memory}\\
\hline
FCN-101    & ResNet-101  & $512\times512$ & 1/8	& $1.04\times10^8$ &   3.25G \\
PSPNet-50 \cite{CVPR17:PSPNET}   & ResNet-50  & $473\times473$  & 1/8	& $0.93\times10^8$  & 1.96G  \\
PSPNet-101 \cite{CVPR17:PSPNET}    & ResNet-101  & $473\times473$   & 1/8	& $1.31\times10^8$ & 3.38G  \\
DeepLab v3+ \cite{TPAMI:DEEPLAB}	& Xception  & $512\times512$   & 1/8	& $0.82\times10^8$  & 5.14G  \\
RefineNet \cite{CVPR17:REFINENET}	& ResNet-101  & $512\times512$ & 1/4	& $2.61\times10^8$ & 1.92G  \\
SANet (ours)	& ResNeXt-50	& $473\times473$ & 1/8	&$1.65\times10^8$    & 2.68G  \\
\hline
\end{tabular}
\end{table*}

\subsection{Computational cost analysis}
 
For deep learning-based models, the computational cost is mainly measured by FLOPs and memory usage of GPU. We compare our SANet based on ResNeXt-50 with some recent deep parsing networks and show the breakdown analysis in Table \ref{TB:FLOPS}. When the input image resolution is $473\times 473$, our model has a moderate computational cost. The parameter efficiency of SANet is similar to PSPNet-101. On the other hand, SANet needs less GPU memory compared to PSPNet with ResNet-101 but a little more float operations. In general, it is worth increasing the FLOPs overhead in a deep parsing network to improve the accuracy of dense prediction.

\section{Experiments}\label{SEC:EXP}

To evaluate the effectiveness of the proposed spatial-semantic aggregation model for scene parsing, we conducted comprehensive experiments on NYU Depth v2 \cite{ECCV12:NYU}, SUN RGB-D \cite{NIPS14:SUN}, Cityscapes \cite{CVPR16:CITYSCAPES} and ADE20K \cite{CVPR17:ADE20K} datasets. In this section, we first briefly introduce the datasets and experimental settings, then report the results on the four datasets.

\subsection{Datasets}

{\bf NYU Depth v2 \& SUN RGB-D}: The two datasets are both indoor scene understanding benchmarks. The pixel-wise classes are from a variety of indoor scenes as recorded by a RGB-D camera from different sensors. We use the standard training/testing splits. The depth images are not used to enhance performance. Some partial experimental results have been shown and analyzed in Section \ref{SUBSEC:MLFA}.

{\bf ADE20K}: This is a challenging dataset with more than 20K scene images, which has 150 classes of dense labels. The testing set has not been released, so we use 2,000 validation images for qualitative evaluation.

{\bf Cityscapes}: This is a street-view dataset taken from 50 European cities, which provides fine-grained pixel-level annotations of 19 classes including buildings, pedestrians, bicycles, cars, etc. The training/validation/testing splits are with 2,975, 500 and 1,525 images, respectively. We do not use the 20,000 coarsely labelled images to pre-train the model.

\subsection{Implementation details}

Our implementation is based on Pytorch. Specifically, we applied the following settings in the experiment: 
\begin{itemize}
    \item To leverage the pixel redundancy and output resolution, we removed the down-sampling operations in the last two learning stages (blocks) of ResNeXt-50 to preserve more visual details without adding extra parameters. Thus, the size of the final feature map is 1/8 of the input image.
    \item We set the dilation rate to 2 for all the $3\times3$ convolutions when the stride is 1 to enlarge the receptive field in the third convolution block. Similarly, we set set the dilation rate to 4 in the forth convolution block. Such a setting is beneficial to improve the segmentation performance without adding computational complexity.
    \item The skip-connection is used multiple times in our SANet. We used the first convolution feature map before residual blocks as the input of the feature aggregation module. Due to the limited computational resource of GPU, we could only use the batch size of 8 at maximum in the training procedure. 
\end{itemize}
We added a pyramid pooling module proposed in \cite{CVPR17:PSPNET} at multi-scale spatial levels to augment the global feature representation prior. To better exploit the contextual information, we employed five bin sizes with $60\times60$, $30\times30$, $20\times20$, $15\times15$ and $10\times10$, respectively. Such a setting is beneficial to parse very small objects or stuff by considering multiple contextual information. After that, a $3\times3$ convolution and an up-sampling were applied to spatially adjust the feature size. 

\subsection{Evaluation metric and experimental settings}

We used {\em intersection-over-union} (IoU) and {\em pixel accuracy} to measure the parsing quality of the models in the experiment. In the training stage, we used horizontal flipping, random scaling and contrast normalization as image augmentation to further improve the model generalization ability. We used the AdamW \cite{ICLR19:ADAMW} optimizer with the initial learning rate $10^{-5}$ and followed \cite{TPAMI:DEEPLAB} to set the learning rate schedualing. In the training process, the best models were checkpointed by the minimum categorical cross-entropy losses on NYU Depth v2 and SUN RGB-D datasets. On ADE20K and Cityscapes datasets, we used the recent lov{\'a}sz-softmax loss \cite{CVPR18:LOVAZ_LOSS} to further optimize the IoU. We applied the multi-scale prediction with a single model in the inference procedure on all datasets, but did not use CRF post-processing \cite{TPAMI:DEEPLAB} to fine-tune the pixel labels after the categorical probability estimation.

\subsection{Results}

\subsubsection{Results on NYU Depth v2 and SUN RGB-D datasets}

\begin{figure}[h]
\centering
    \includegraphics[width=0.4\textwidth]{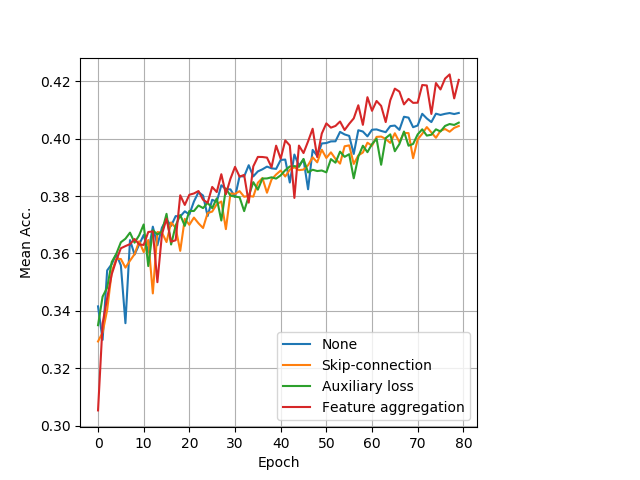}   
    \caption{The mean accuracy curves on the NYU Depth v2 validation set. The feature aggregation is based on a dilated FCN. Due to the multiple skip-connections, the proposed feature aggregation can effectively accelerate the training of the deep scene parsing model.}
\label{FIG:CONV}
\end{figure}

One of the nice properties of the proposed multi-layer feature aggregation is that the gradient can be simultaneously passed to multiple layers, making the deep parsing network easy to train. In the model training of the FCN on NYU Depth v2 dataset, we recorded the mean accuracy values for the first 80 training epochs on the validation set, which is shown in Fig. \ref{FIG:CONV}. Compared to the FCN models with skip-connections and auxiliary losses, the proposed feature aggregation module can simultaneously pass the gradients into multiple learning blocks, including the shallower convolution blocks, which improves the optimization efficiency for dense predictions.  

We first conducted the component analysis of SANet on NYU Depth v2 dataset. The proposed SANet is essentially a composition of a stem network (ResNeXt-50), a multi-layer feature aggregation module and a pyramid pooling module. In the training process, the best models were checkpointed by the minimum categorical cross-entropies, and the final evaluation of the testing set is summarized in Table \ref{TB:COMP}. Applying a pyramid pooling module on a FCN can slightly improve the parsing performance. If the stem network is augmented by the feature aggregation module, the accuracy can be further boosted. Specifically, the much lower loss value (categorical cross-entropy in our case) means the model can get more reliable parsing results.


\begin{table}[h]
\centering
\caption{Component analysis of SANet on NYU Depth v2 testing set.}
\label{TB:COMP}
\begin{tabular}{|c|c|c|c|}
\hline
{\bf Method}	&PSP & Feature aggregation &Pixel accuracy   \\
\hline
ResNet-50 & & &71.8  \\
ResNeXt-50 & & &72.5   \\
ResNeXt-50 &\checkmark & &74.3   \\
ResNeXt-50 &\checkmark &\checkmark &{\bf 75.9}   \\
\hline
\end{tabular}
\end{table}

\begin{table}[h]
\centering
\caption{Scene parsing results on NYU Depth v2 and SUN RGB-D testing sets.}
\label{TB:NYU-SUN}
\begin{tabular}{|c|cc|cc|}
\hline
\multirow{2}{*}{Methods} & \multicolumn{2}{c|}{NYU Depth v2} & \multicolumn{2}{c|}{SUN RGB-D}	 \\
\cline{2-5}  &Pixel Acc. &IoU  &Pixel Acc. &IoU  \\
\hline
SegNet \cite{TPAMI:SEGNET}  &- &-  &72.6 &31.8 \\
SEGCloud \cite{IC3DV17:SEFCLOUD} &- &43.5  &- &-   \\
Lin et al. \cite{CVPR16:PIECEWISE}	&70.0 &40.6  &78.4 &42.3   \\
RefineNet\cite{CVPR17:REFINENET}	&74.4 &47.6  &81.1 &47.0   \\
MSCI \cite{ECCV18:MCCI}		 &- &49.0 &- &50.4 \\
Pad-Net \cite{CVPR18:PAD_NET} &75.2 &50.2 &- &- \\
CCL \cite{CVPR18:CCL} &- &- &81.4 &47.1 \\
\hline
SANet (ours)	&{\bf 75.9} &{\bf 50.7}  &{\bf 82.3} &{\bf 51.5}   \\
\hline
\end{tabular}
\end{table}

We show the quantitative results of the two indoor scene parsing datasets in Table \ref{TB:NYU-SUN}. Even we did not incorporate the depth images in the training process, the proposed SANet reaches the best performance in most cases. Since we adopted the pyramid pooling module which is the same as PSPNet, the proposed spatial-semantic feature aggregation module outperforms the pixel accuracies by 0.7\% and 0.9\%, and boosts the IoU values by 0.5\% and 1.1\% on the two datasets, respectively.

\subsubsection{Results on ADE20K dataset}

\begin{table}[t]
\centering
\caption{Results on ADE20K validation set.}
\label{TB:ADE20K}
\begin{tabular}{|c|cc|}
\hline
{\bf Method}	&Pixel Acc. &IoU  \\
\hline
PSPNet\cite{CVPR17:PSPNET}  &81.4 &43.3 \\
SAC\cite{ICCV17:SAC}  &81.9 &44.3 \\
RefineNet\cite{CVPR17:REFINENET}  &- &42.4  \\
PSANet\cite{ECCV18:PSANET}  &81.5 &43.8 \\
EncNet\cite{CVPR18:ENCNET} & 81.7 & 44.7 \\
\hline
SANet (ours) &{\bf 82.1} &{\bf 44.8}   \\
\hline
\end{tabular}
\end{table}

\begin{figure*}[t]\centering
\begin{minipage}{0.2\textwidth}
	\includegraphics[width=1\textwidth]{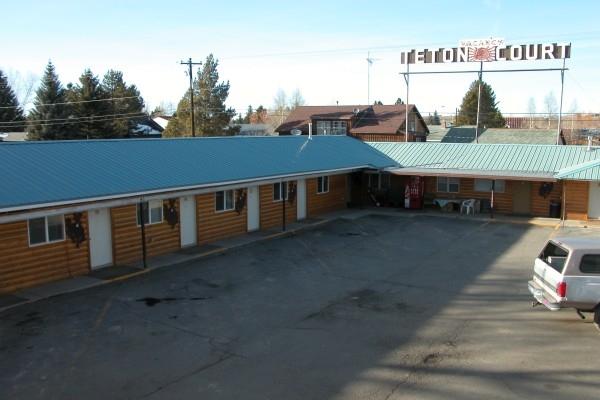}
\end{minipage}
\begin{minipage}{0.2\textwidth}
	\includegraphics[width=1\textwidth]{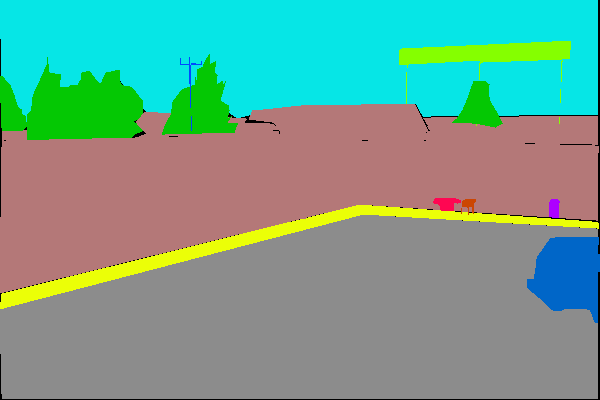}
\end{minipage}	
\begin{minipage}{0.2\textwidth}	
	\includegraphics[width=1\textwidth]{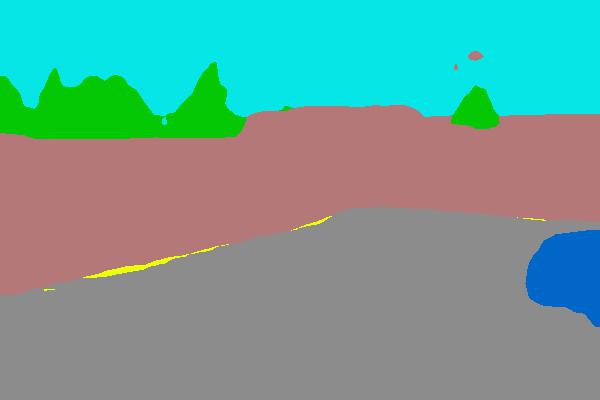}    	
\end{minipage}
\begin{minipage}{0.2\textwidth}
	\includegraphics[width=1\textwidth]{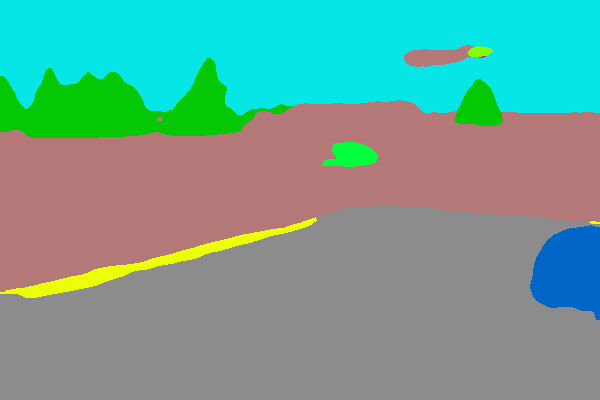}
\end{minipage}

\begin{minipage}{0.2\textwidth}
	\includegraphics[width=1\textwidth]{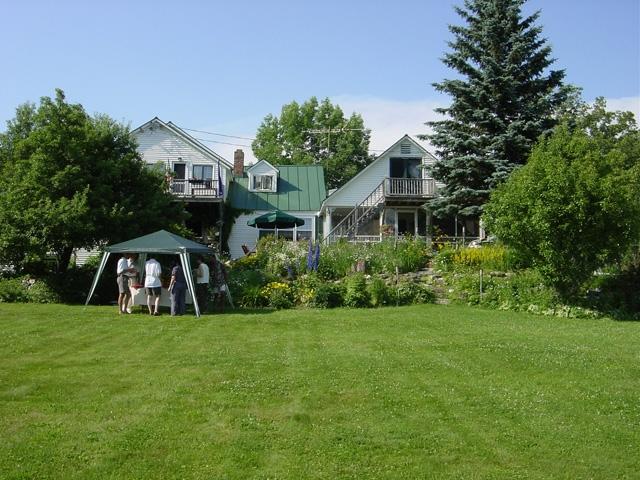}
\end{minipage}
\begin{minipage}{0.2\textwidth}
	\includegraphics[width=1\textwidth]{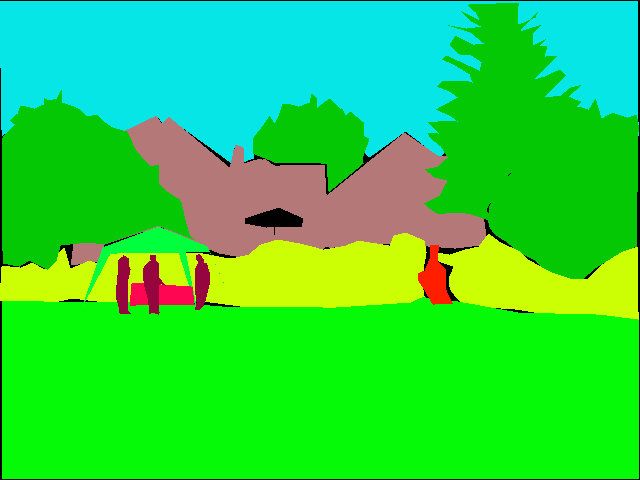}
\end{minipage}
\begin{minipage}{0.2\textwidth}	
	\includegraphics[width=1\textwidth]{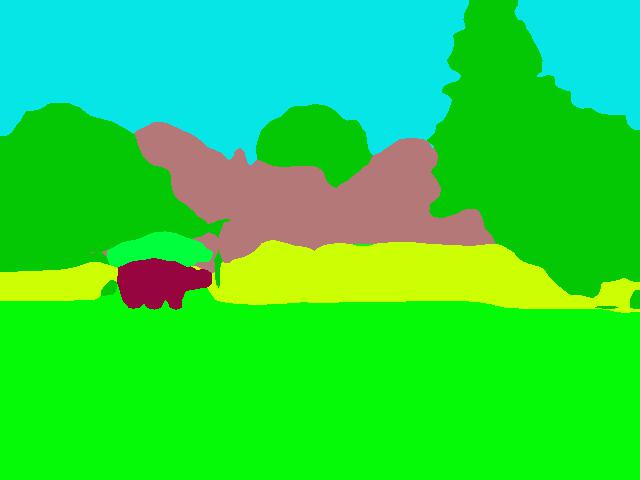}    	
\end{minipage}
\begin{minipage}{0.2\textwidth}
	\includegraphics[width=1\textwidth]{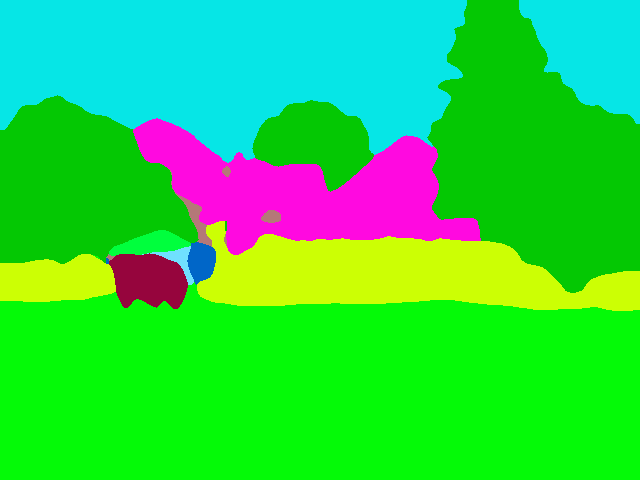}
\end{minipage}

\begin{minipage}{0.2\textwidth}
	\includegraphics[width=1\textwidth]{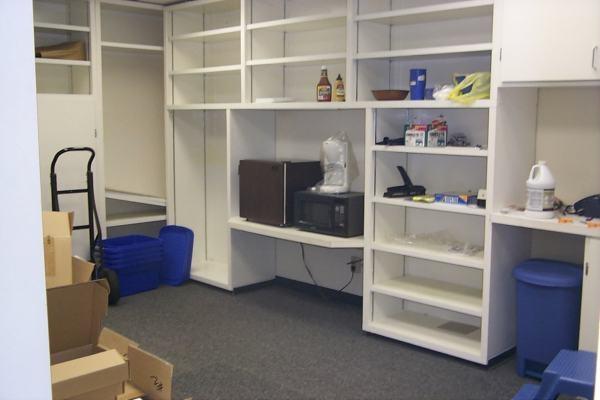}
\end{minipage}
\begin{minipage}{0.2\textwidth}
	\includegraphics[width=1\textwidth]{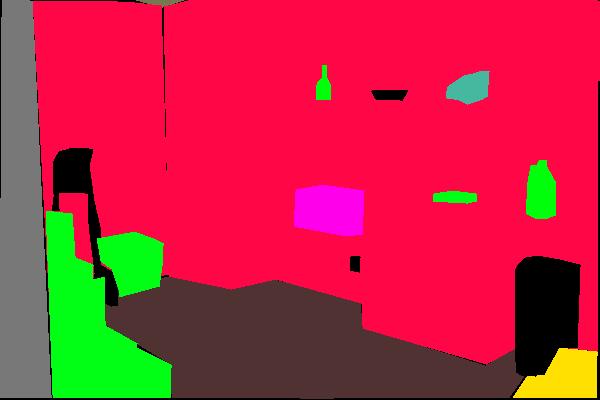}
\end{minipage}
\begin{minipage}{0.2\textwidth}	
	\includegraphics[width=1\textwidth]{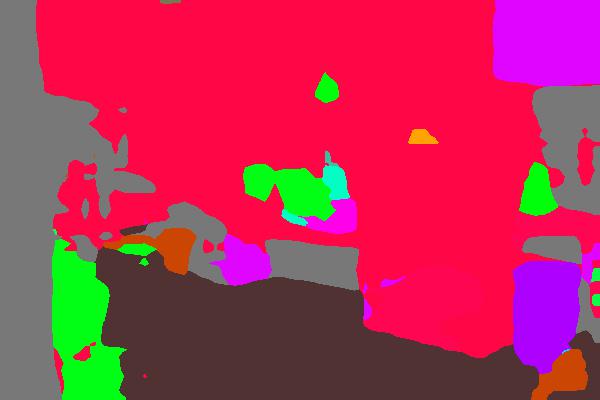}    	
\end{minipage}
\begin{minipage}{0.2\textwidth}
	\includegraphics[width=1\textwidth]{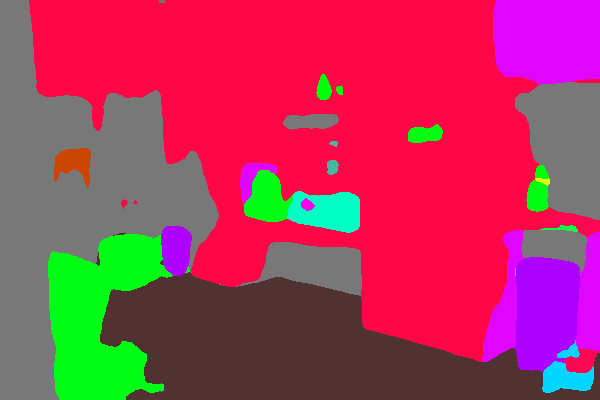}
\end{minipage}

\begin{minipage}{0.2\textwidth}\centering
	\includegraphics[width=1\textwidth]{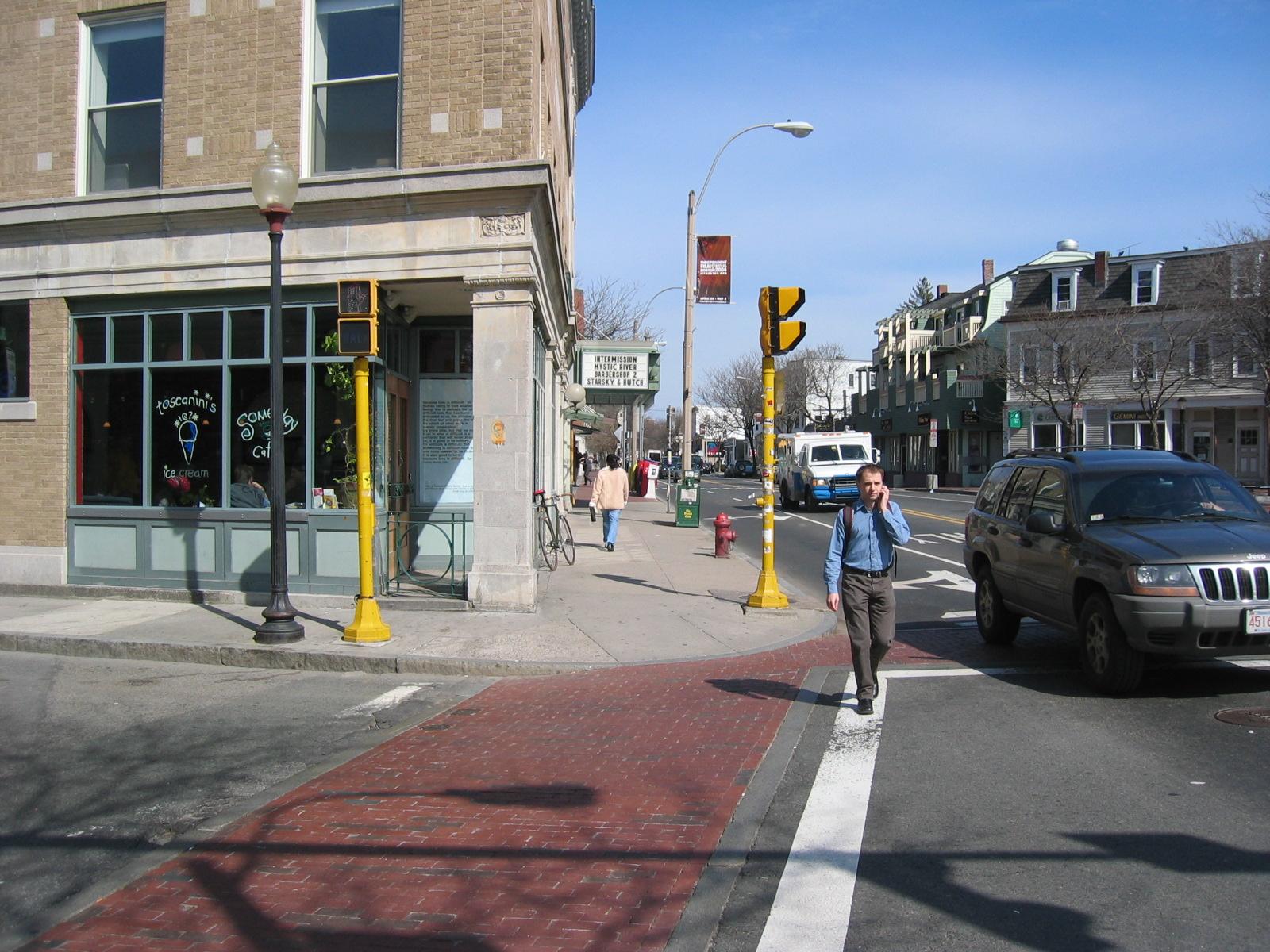}
(a) Image 
\end{minipage}	
\begin{minipage}{0.2\textwidth}\centering	
	\includegraphics[width=1\textwidth]{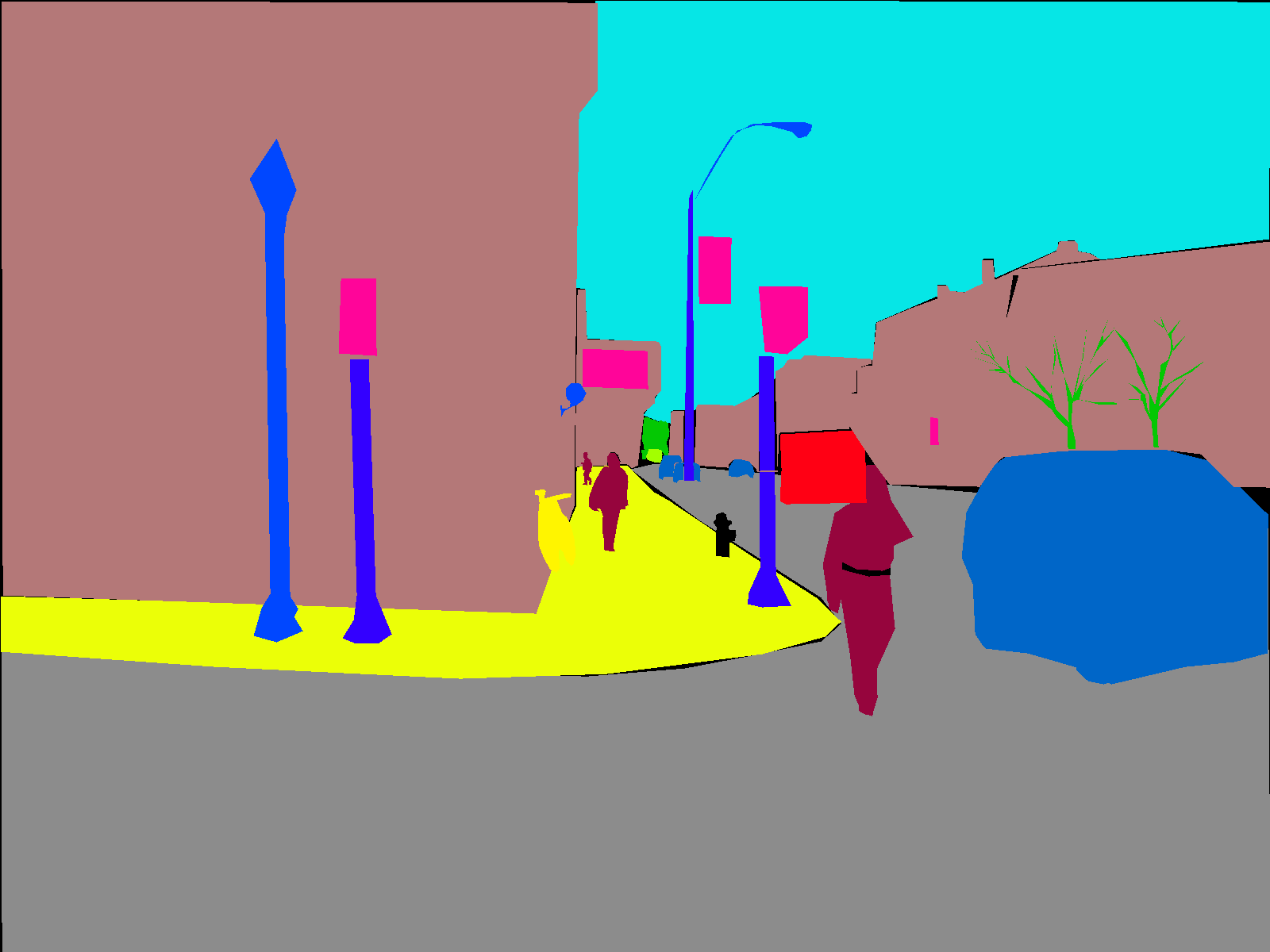}
(b) GT 
\end{minipage}
\begin{minipage}{0.2\textwidth}\centering	
	\includegraphics[width=1\textwidth]{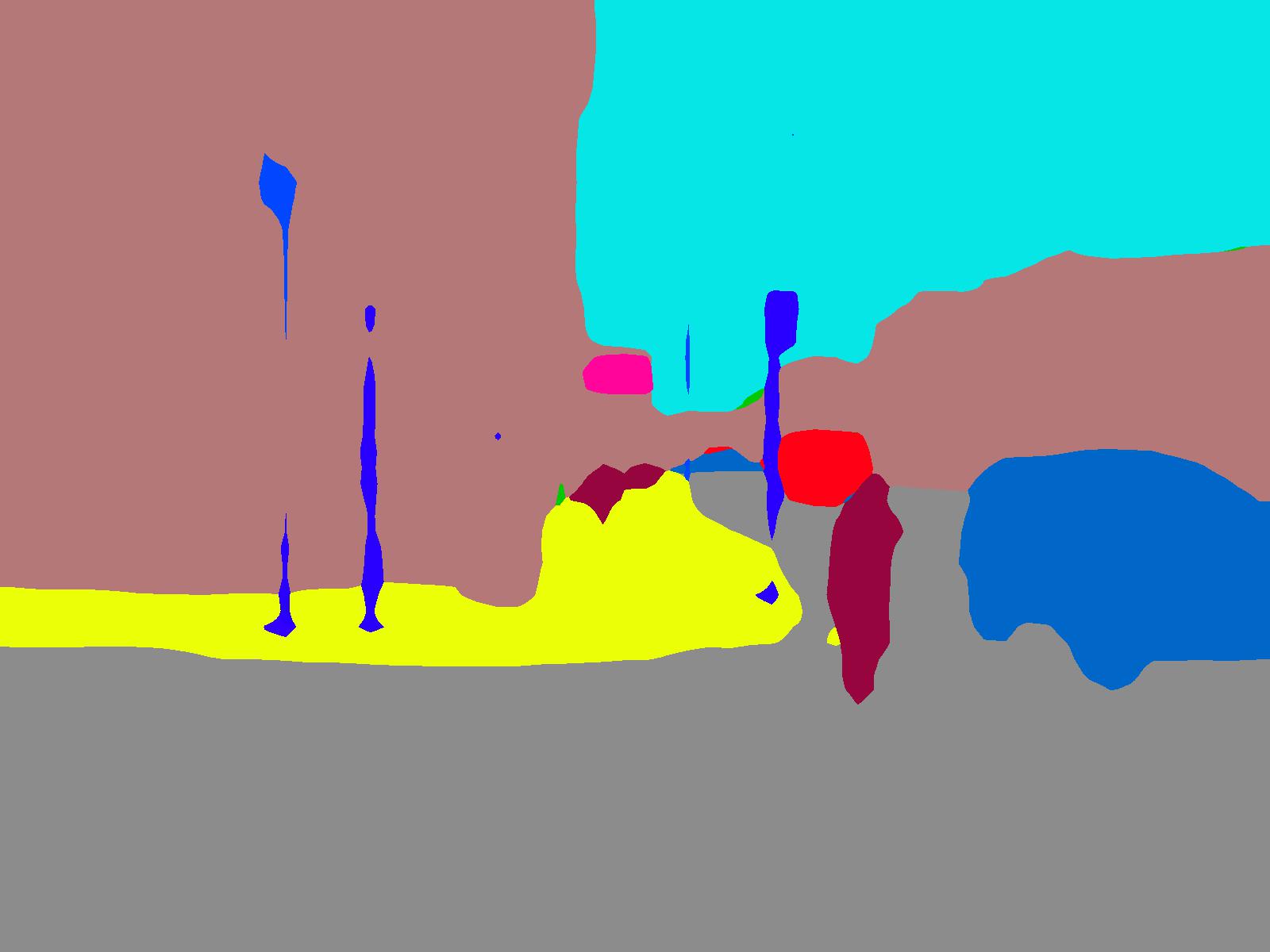} 
(c) PSPNet
\end{minipage}
\begin{minipage}{0.2\textwidth}\centering	
	\includegraphics[width=1\textwidth]{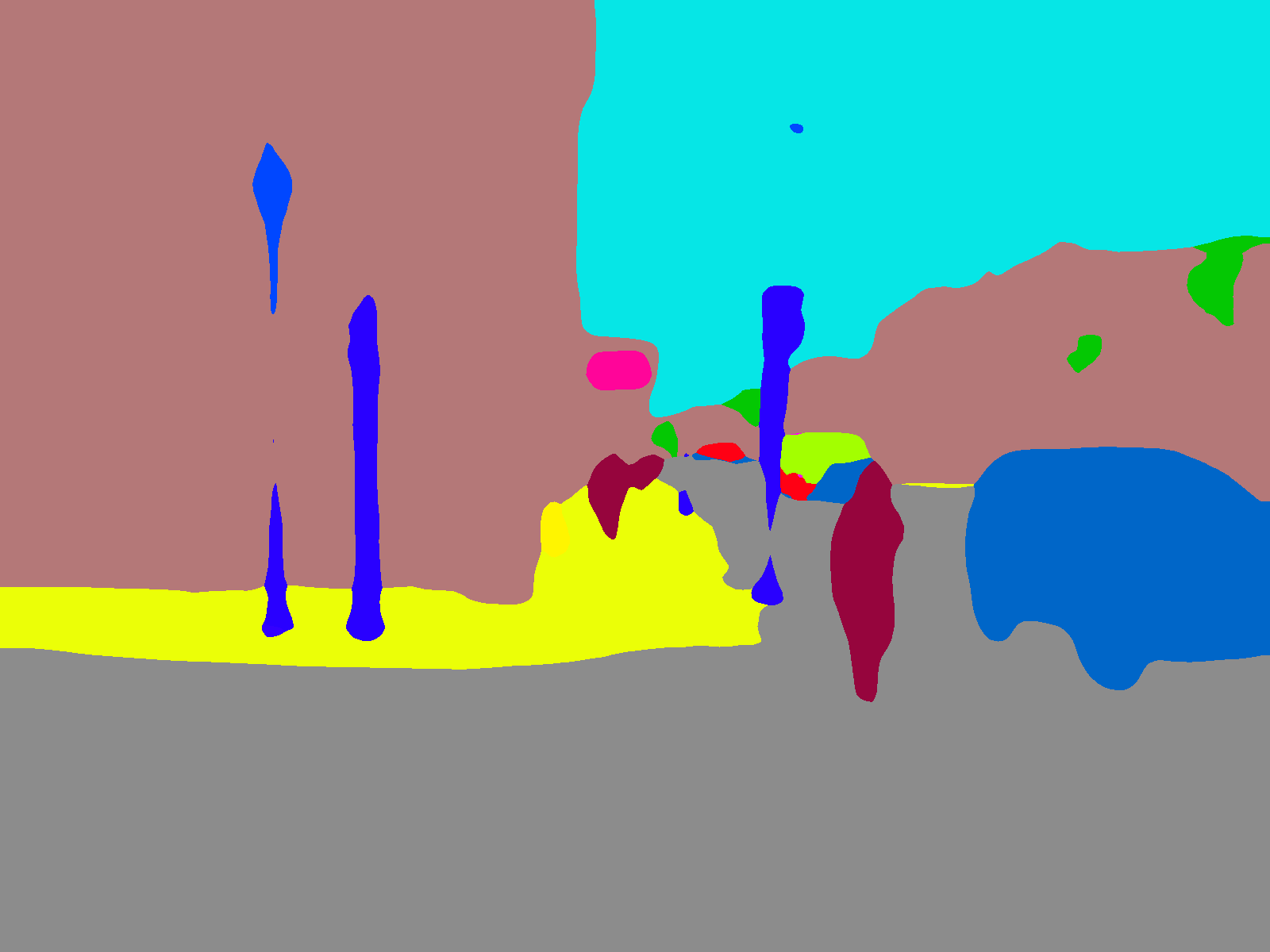}
(d) SANet 
\end{minipage}	
\caption{Scene parsing examples on ADE20K validation set.}
\label{FIG:ADE20K}
\end{figure*}

We experimented on the large-scale ADE20K dataset to verify the effectiveness of the proposed SANet. The comparisons on the validation set with some recently proposed methods are reported in Table \ref{TB:ADE20K}. Results show that our model achieves 82.1\% in pixel accuracy and 44.7\% in IoU, which achieves the best performance among these methods. Some example scene parsing results in both indoor and outdoor environments are illustrated in Fig. \ref{FIG:ADE20K}.

\subsubsection{Results on Cityscapes dataset}

\begin{figure*}[t]\centering
\begin{minipage}{0.23\textwidth}
	\includegraphics[width=1\textwidth]{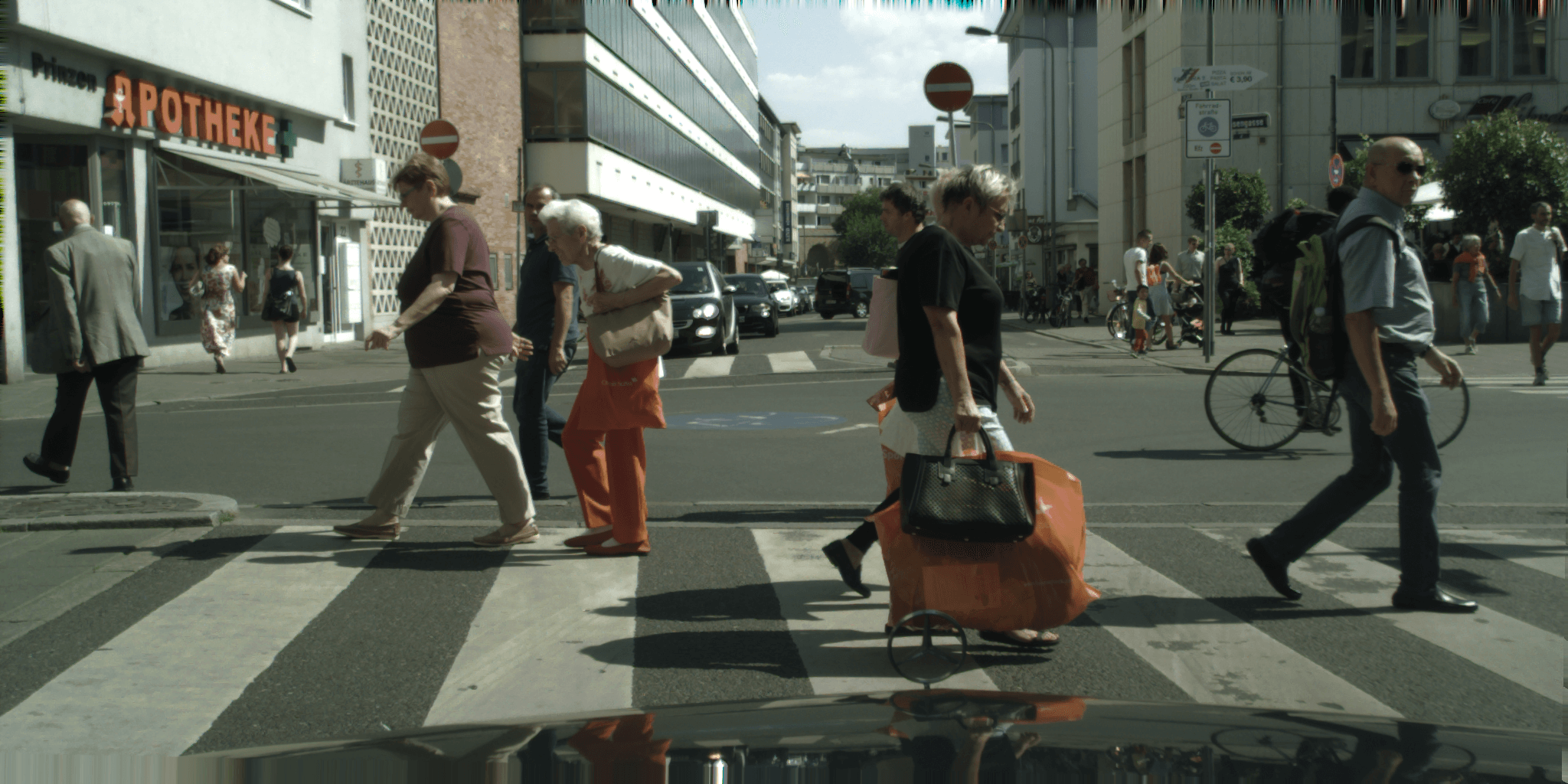}
\end{minipage}
\begin{minipage}{0.23\textwidth}
	\includegraphics[width=1\textwidth]{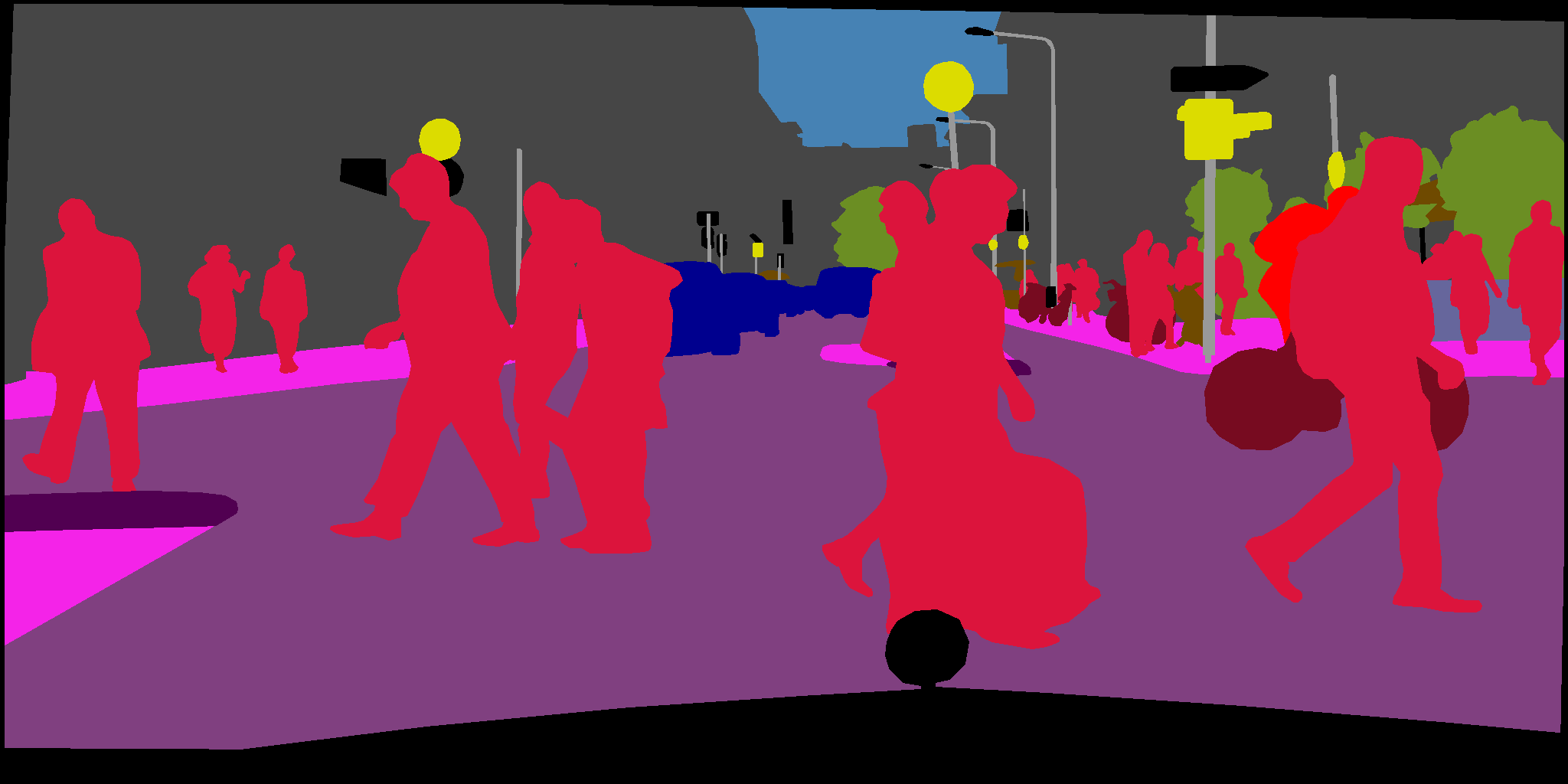}
\end{minipage}	
\begin{minipage}{0.23\textwidth}	
	\includegraphics[width=1\textwidth]{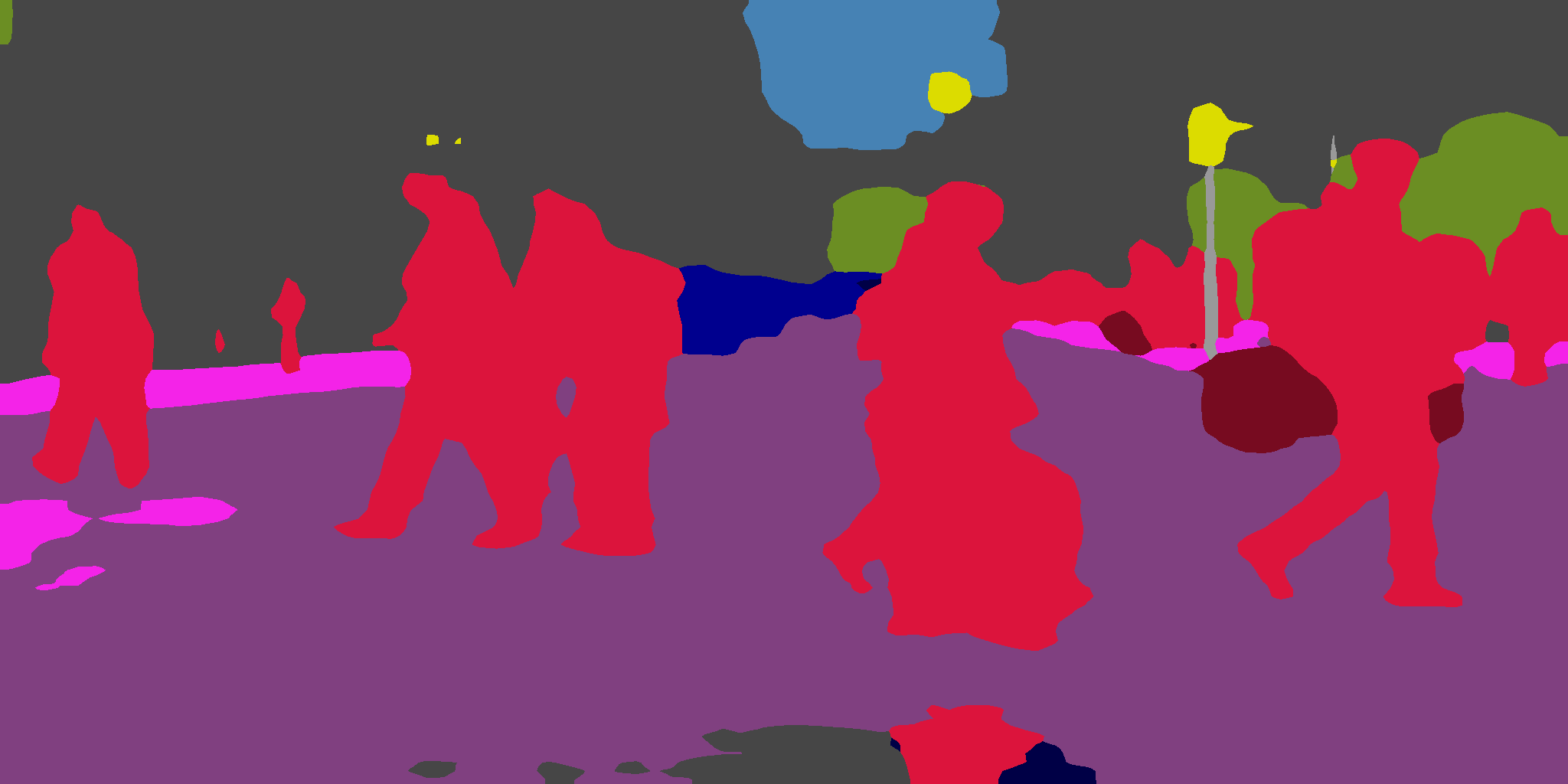}    	
\end{minipage}
\begin{minipage}{0.23\textwidth}
	\includegraphics[width=1\textwidth]{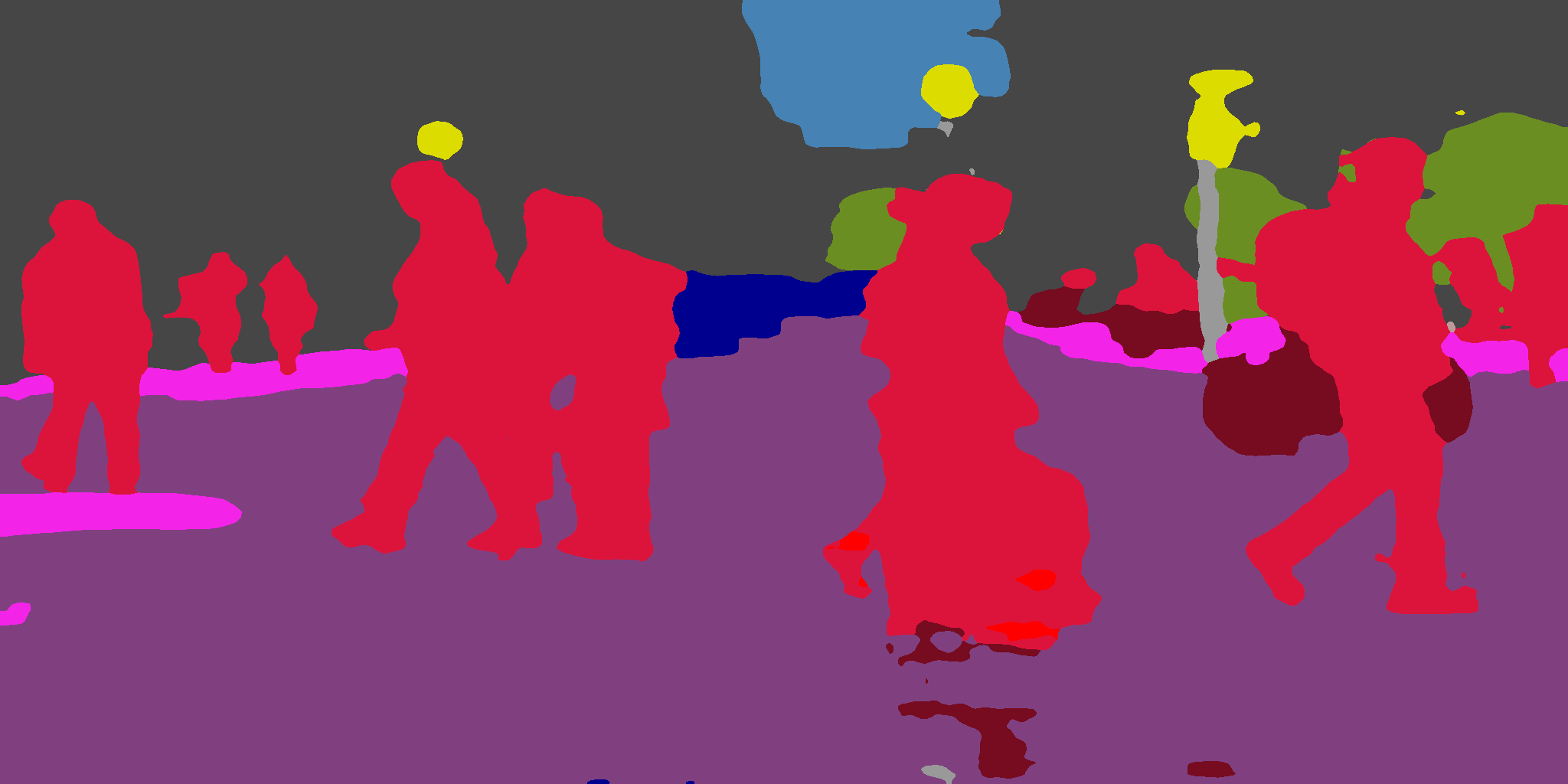}
\end{minipage}

\begin{minipage}{0.23\textwidth}
	\includegraphics[width=1\textwidth]{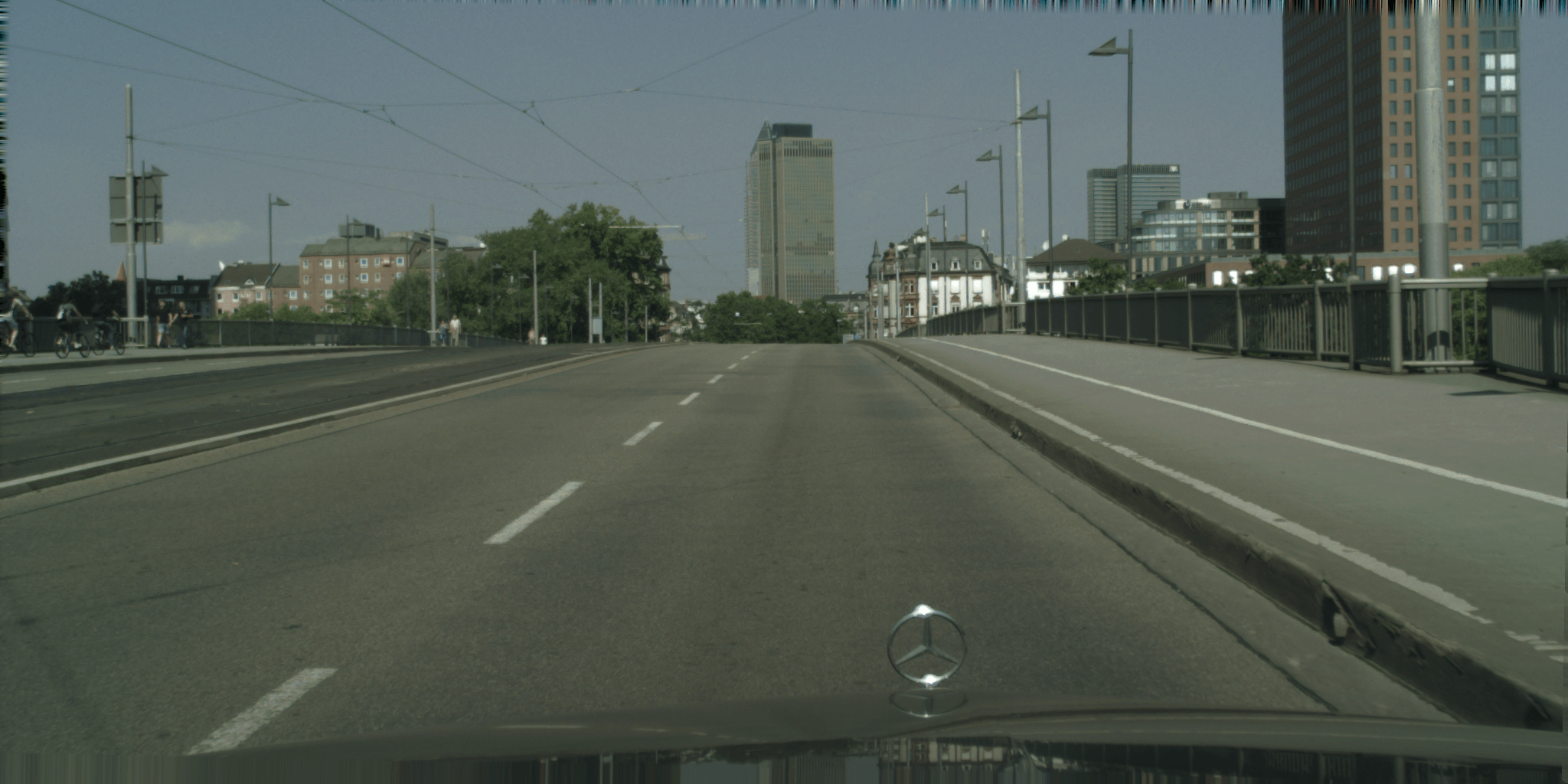}
\end{minipage}
\begin{minipage}{0.23\textwidth}
	\includegraphics[width=1\textwidth]{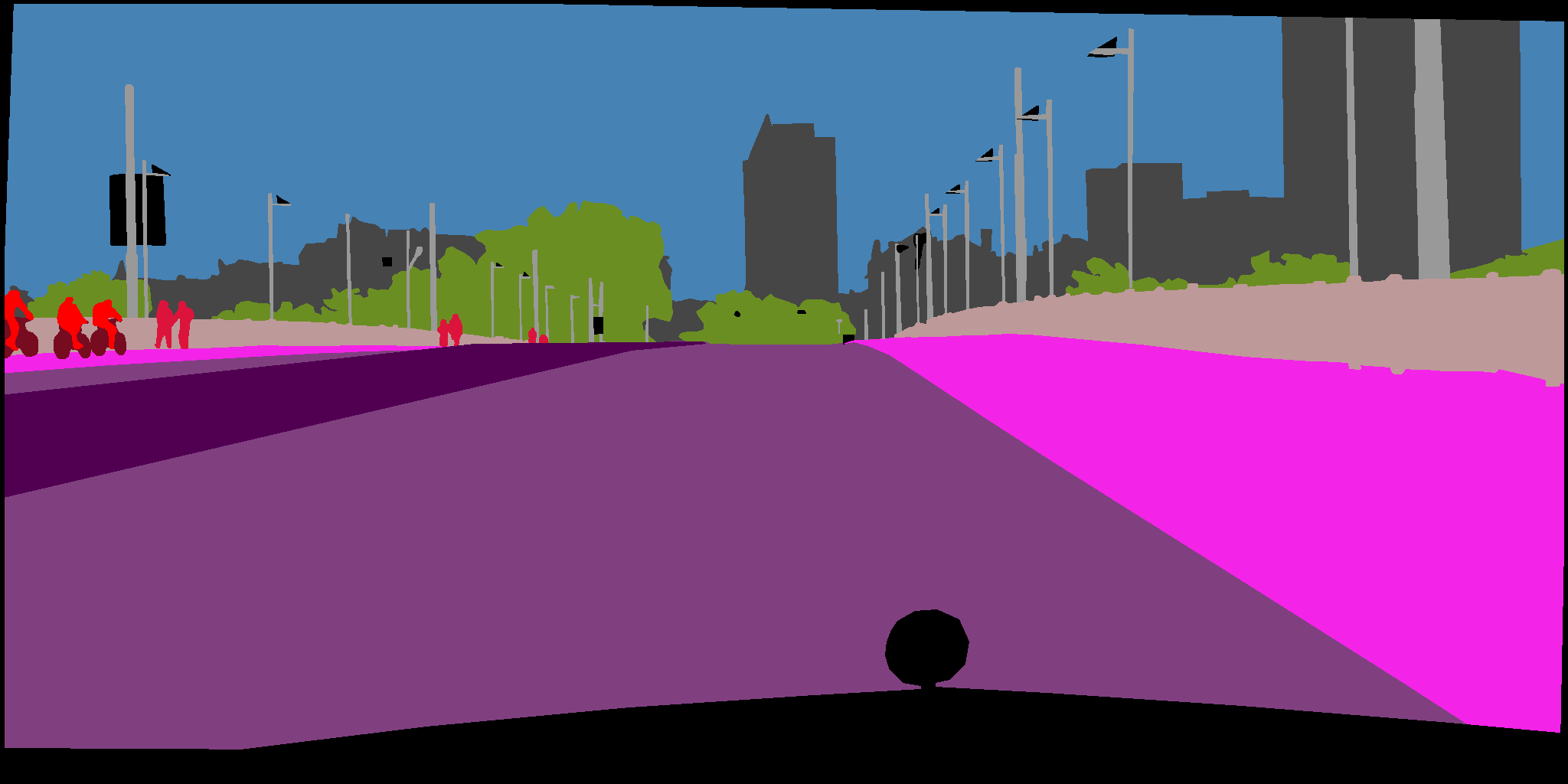}
\end{minipage}
\begin{minipage}{0.23\textwidth}	
	\includegraphics[width=1\textwidth]{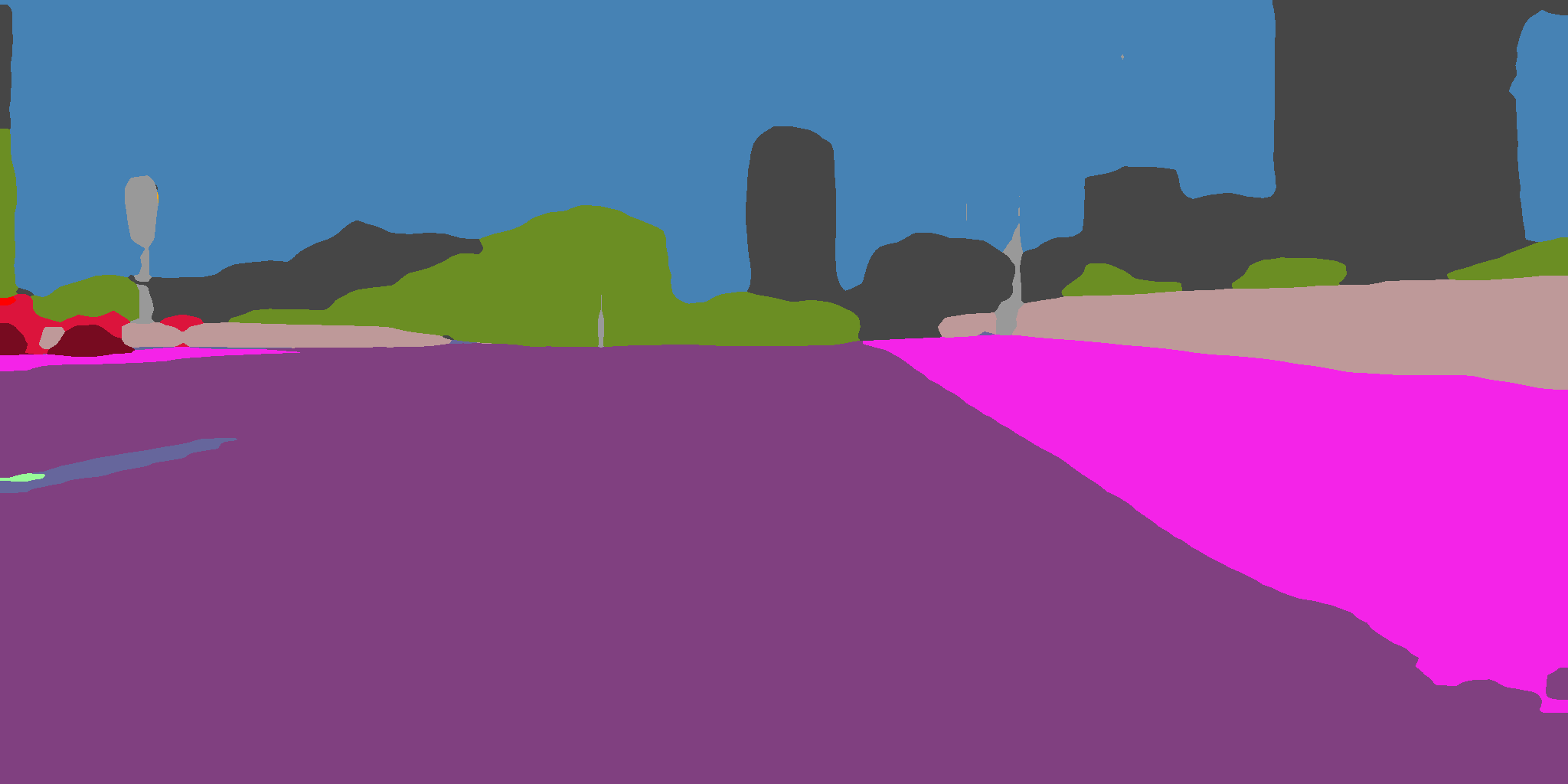}    	
\end{minipage}
\begin{minipage}{0.23\textwidth}
	\includegraphics[width=1\textwidth]{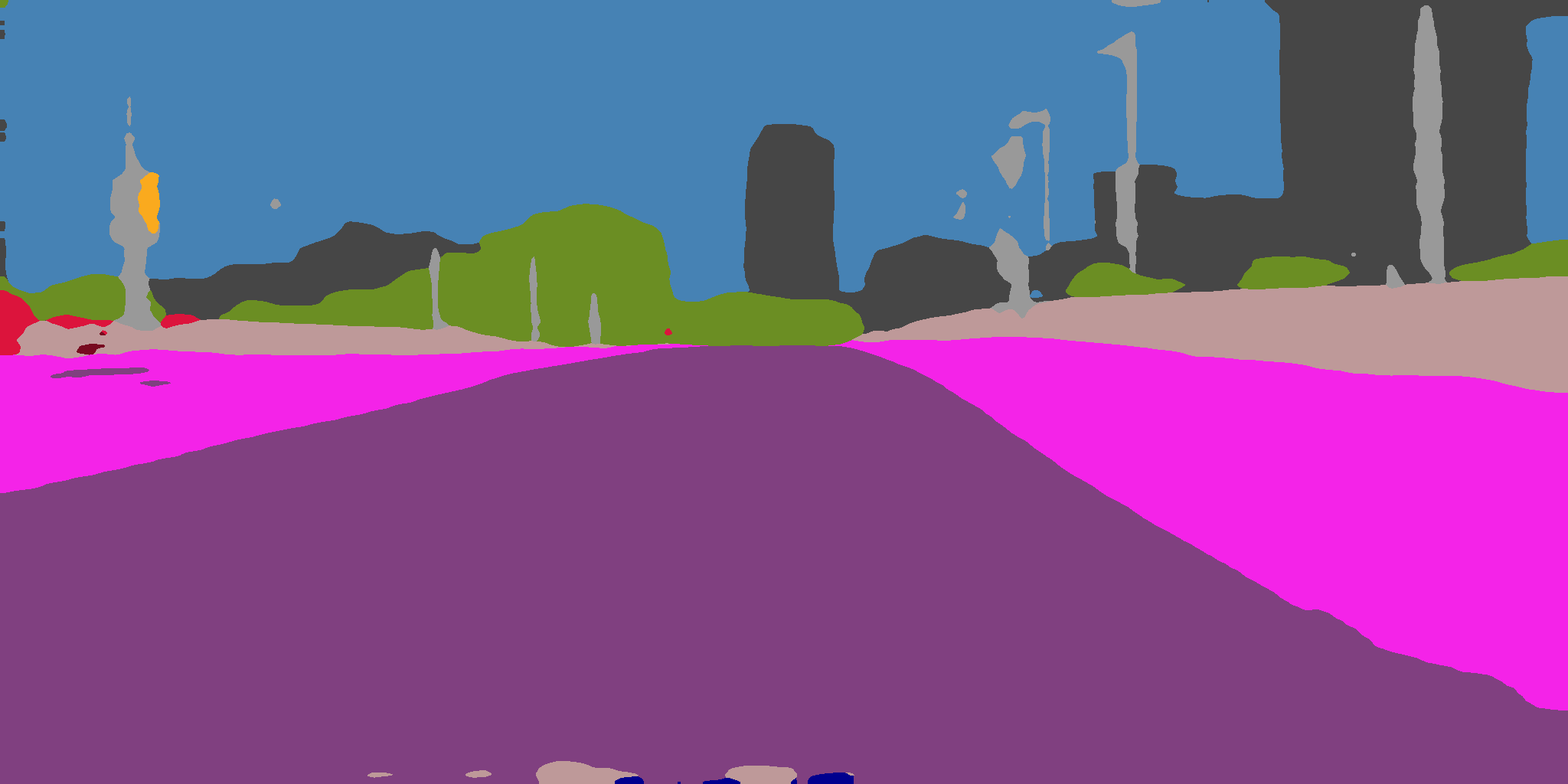}
\end{minipage}

\begin{minipage}{0.23\textwidth}
	\includegraphics[width=1\textwidth]{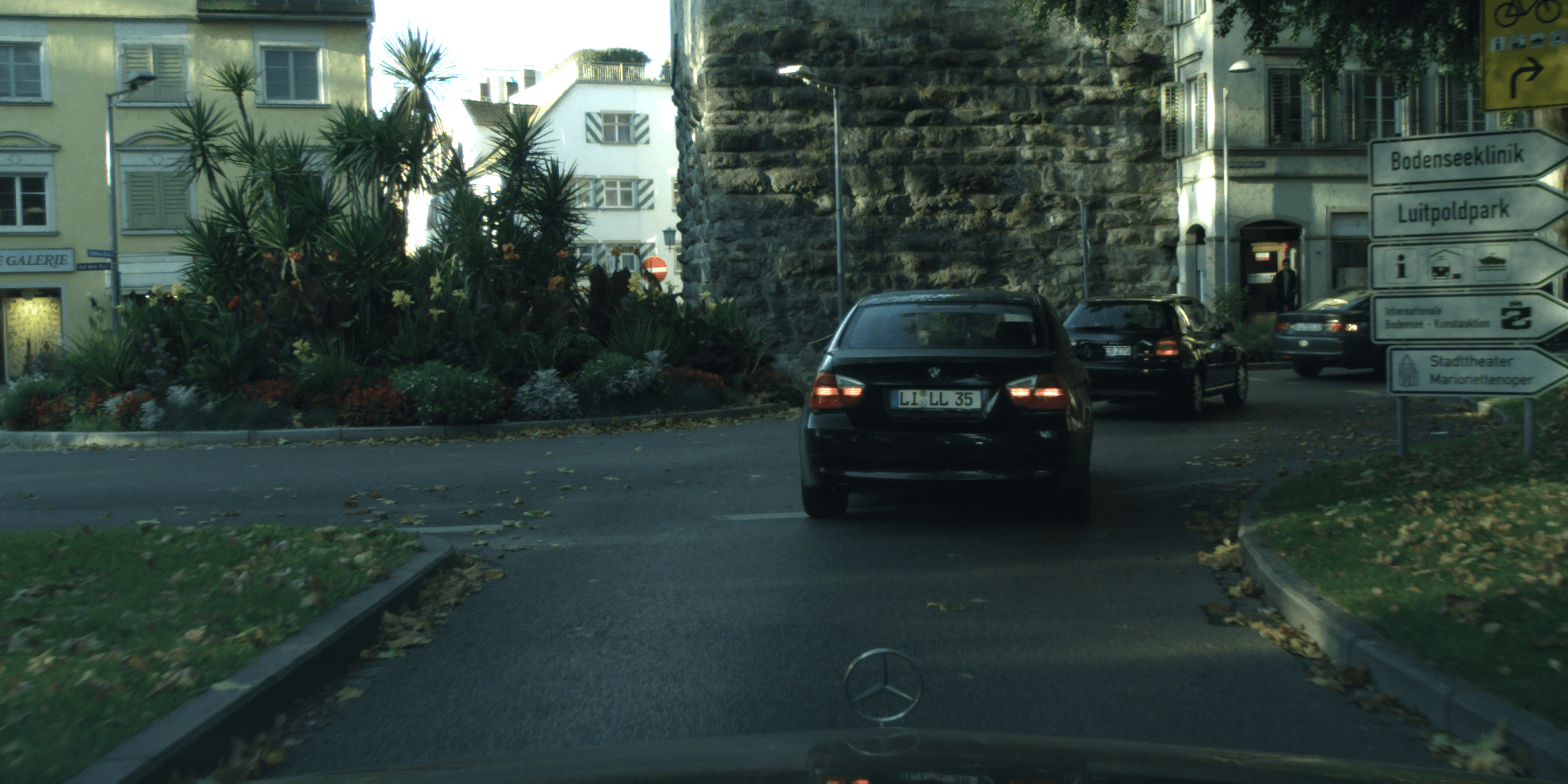}
\end{minipage}
\begin{minipage}{0.23\textwidth}
	\includegraphics[width=1\textwidth]{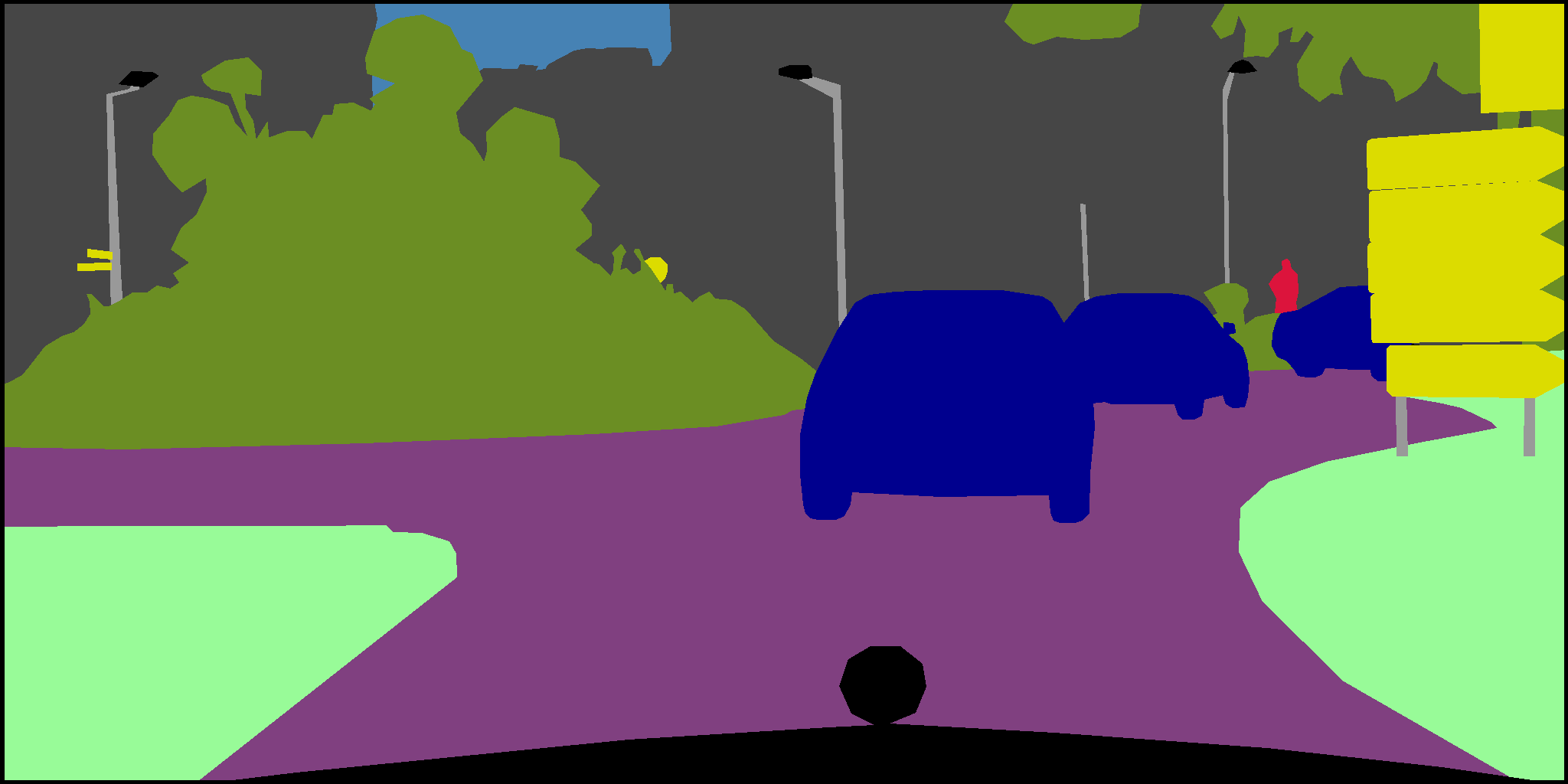}
\end{minipage}
\begin{minipage}{0.23\textwidth}	
	\includegraphics[width=1\textwidth]{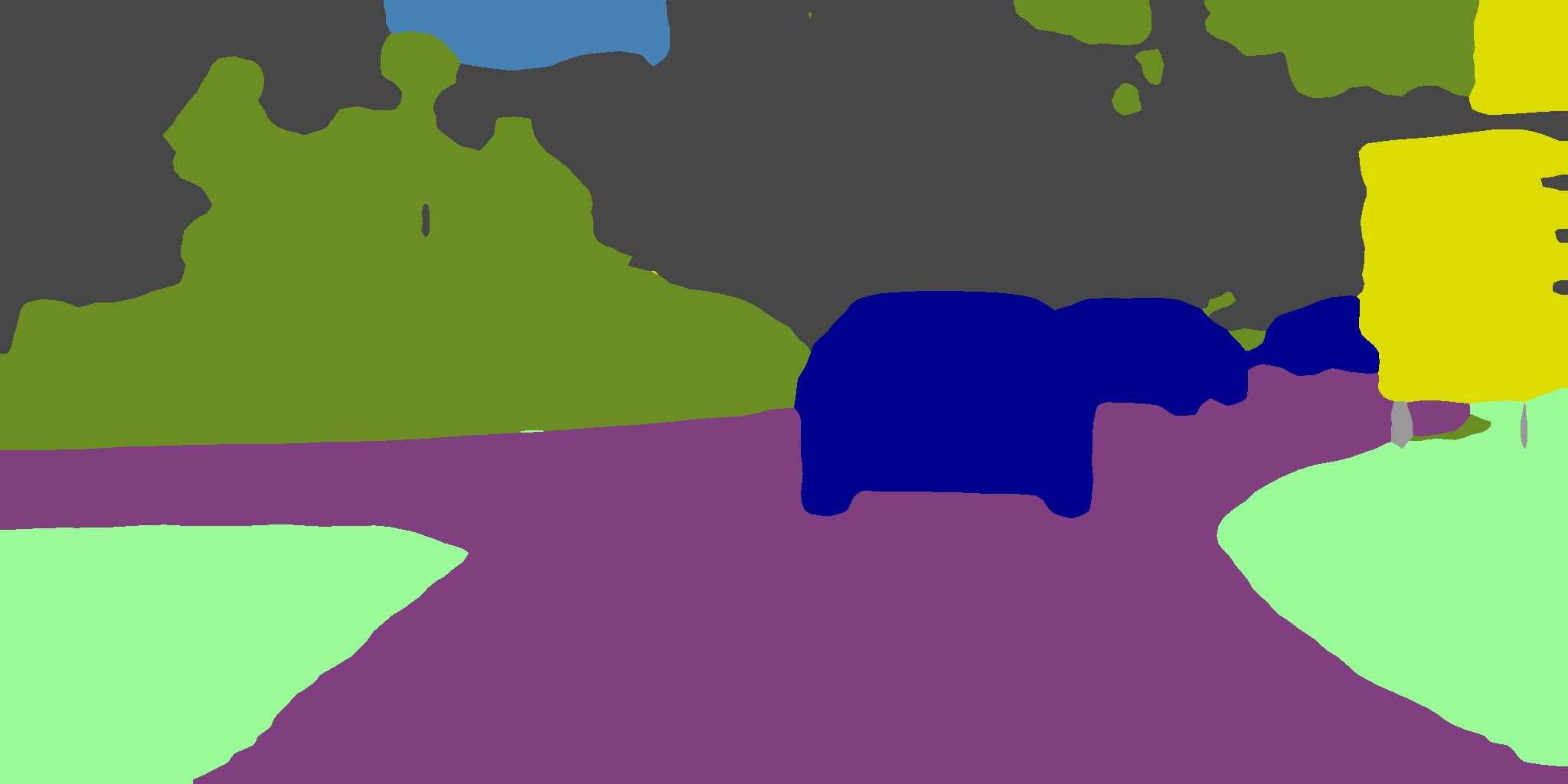}    	
\end{minipage}
\begin{minipage}{0.23\textwidth}
	\includegraphics[width=1\textwidth]{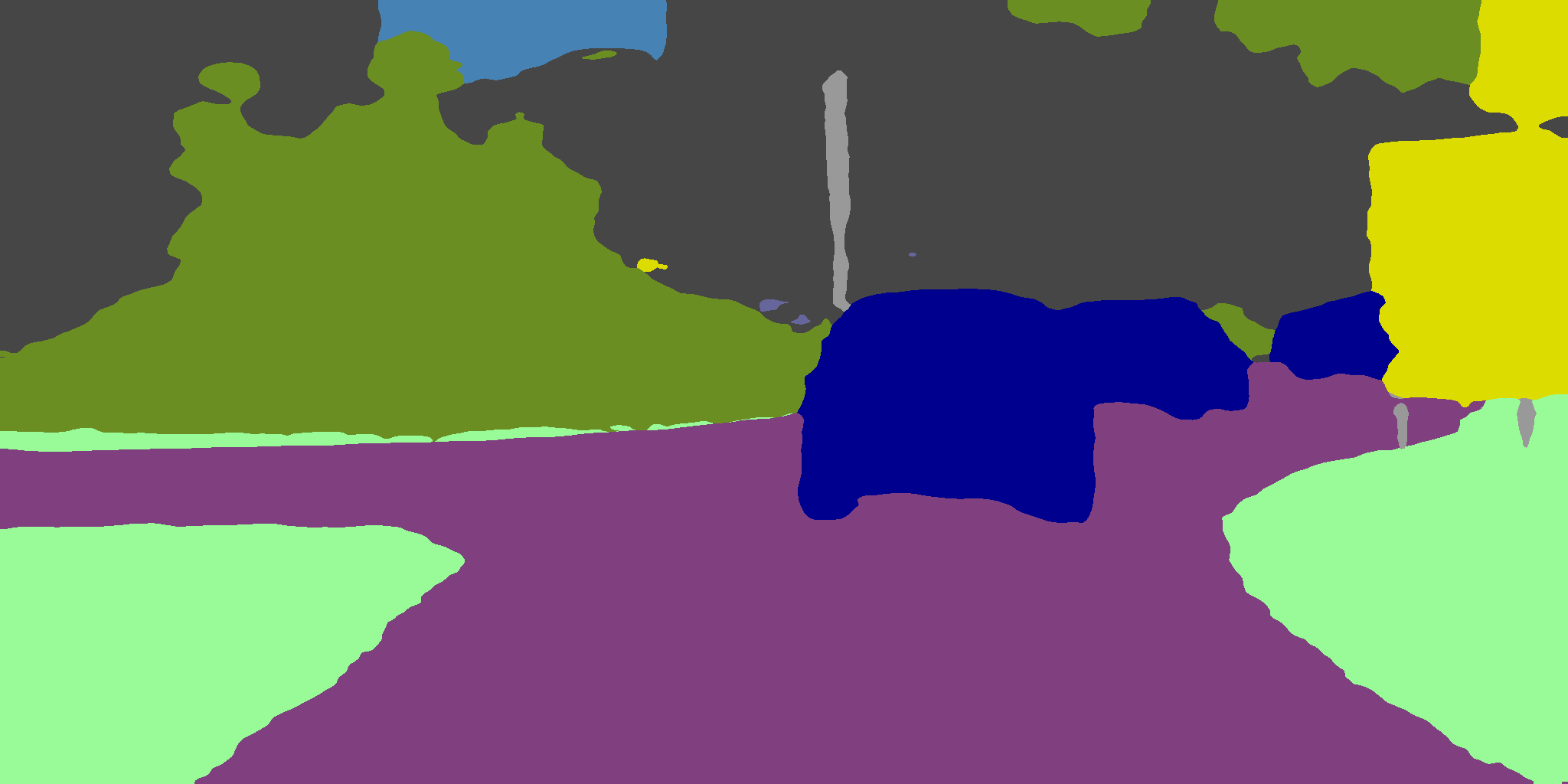}
\end{minipage}

\begin{minipage}{0.23\textwidth}\centering
	\includegraphics[width=1\textwidth]{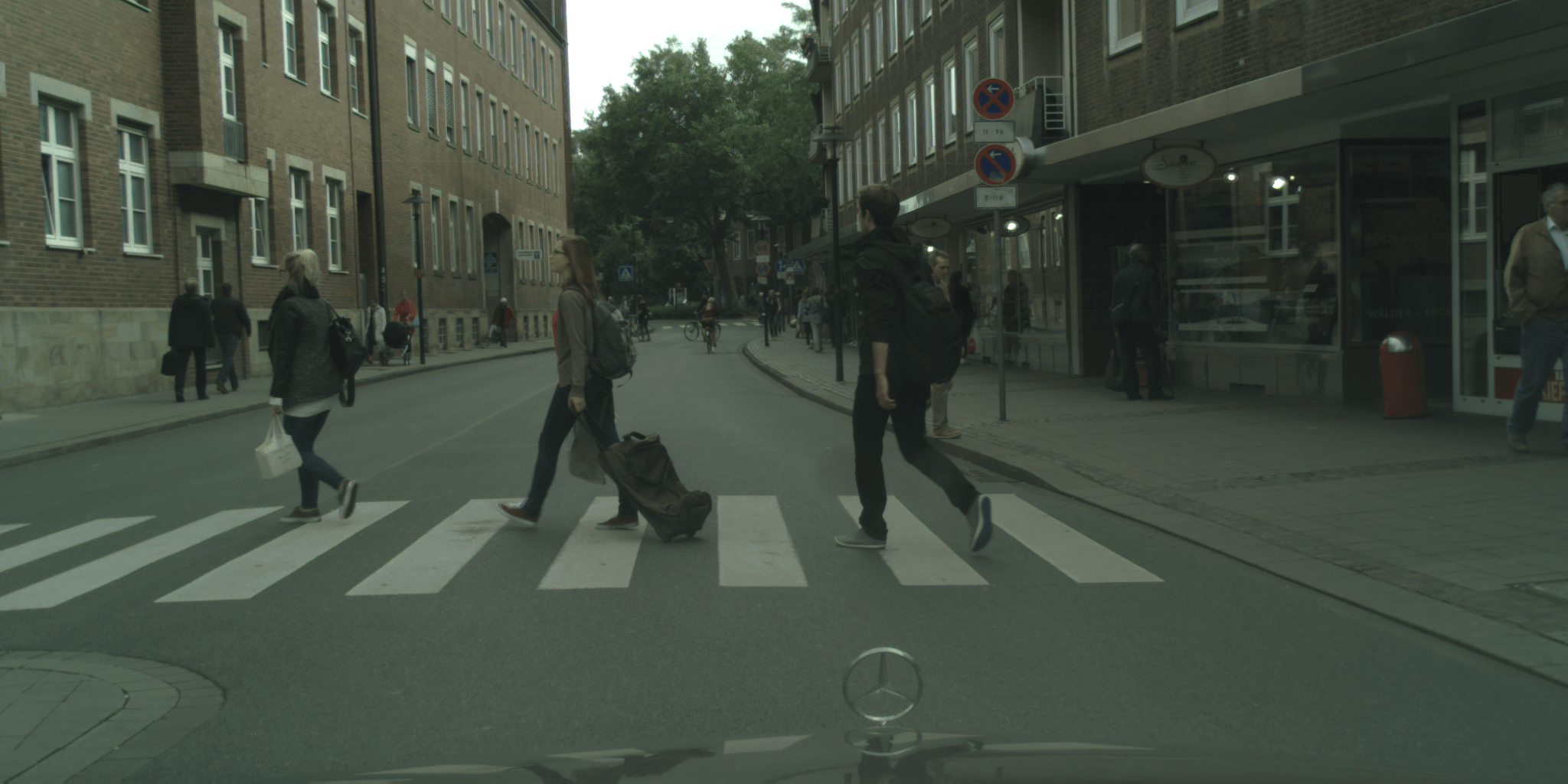}
(a) Image 
\end{minipage}	
\begin{minipage}{0.23\textwidth}\centering
	\includegraphics[width=1\textwidth]{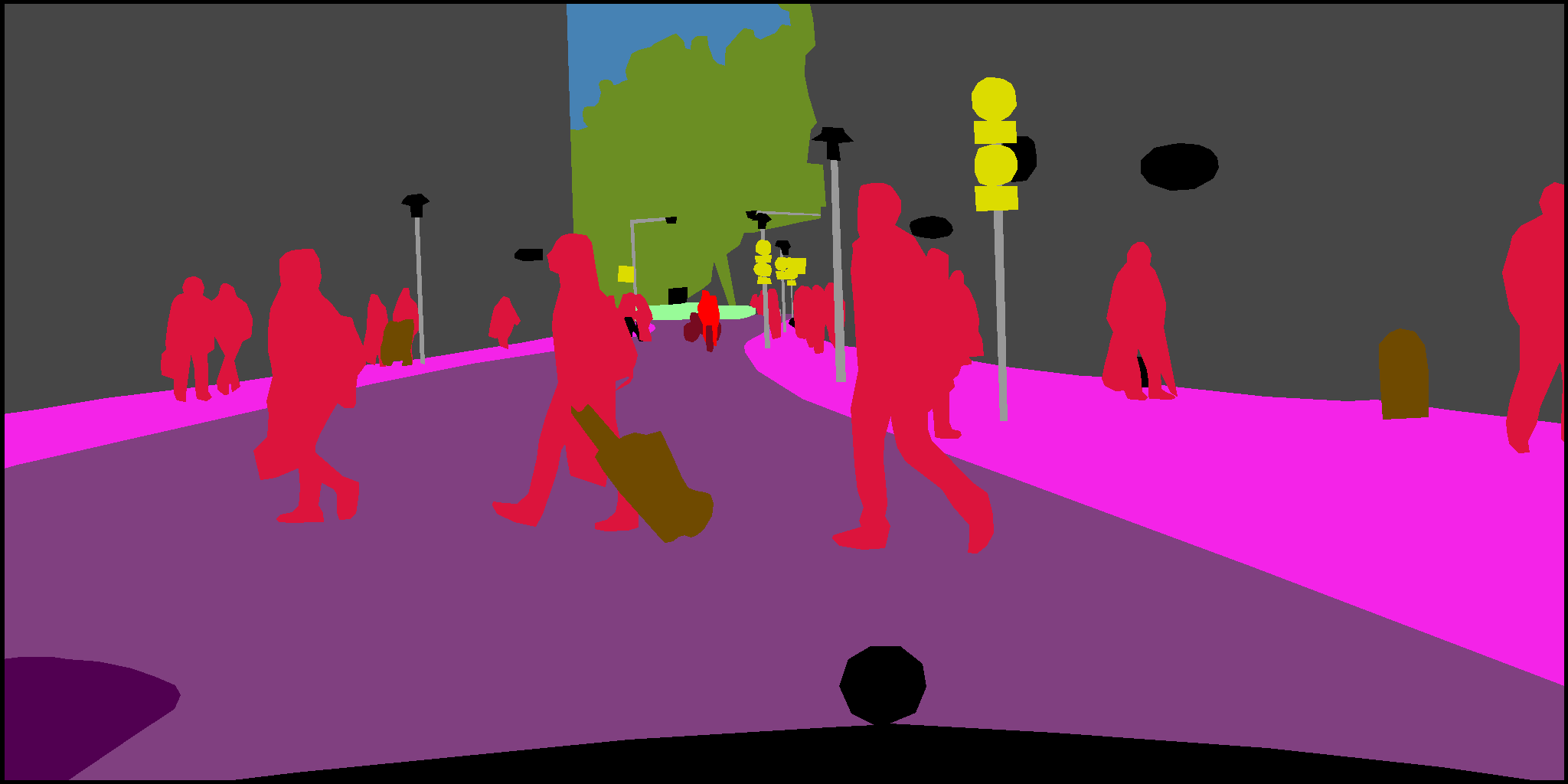}
(b) GT
\end{minipage}
\begin{minipage}{0.23\textwidth}\centering	
	\includegraphics[width=1\textwidth]{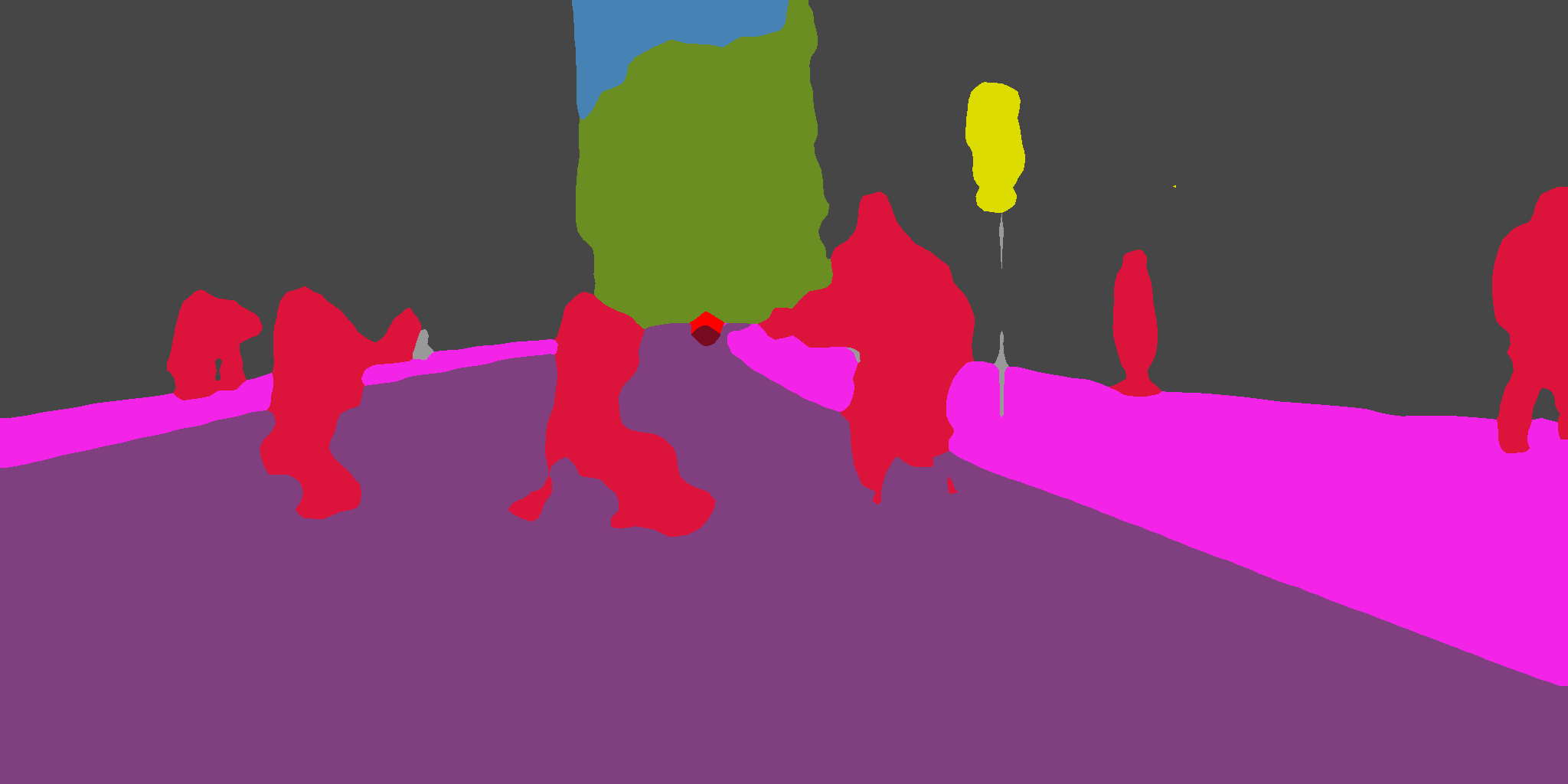} 
(c) PSPNet
\end{minipage}
\begin{minipage}{0.23\textwidth}\centering	
	\includegraphics[width=1\textwidth]{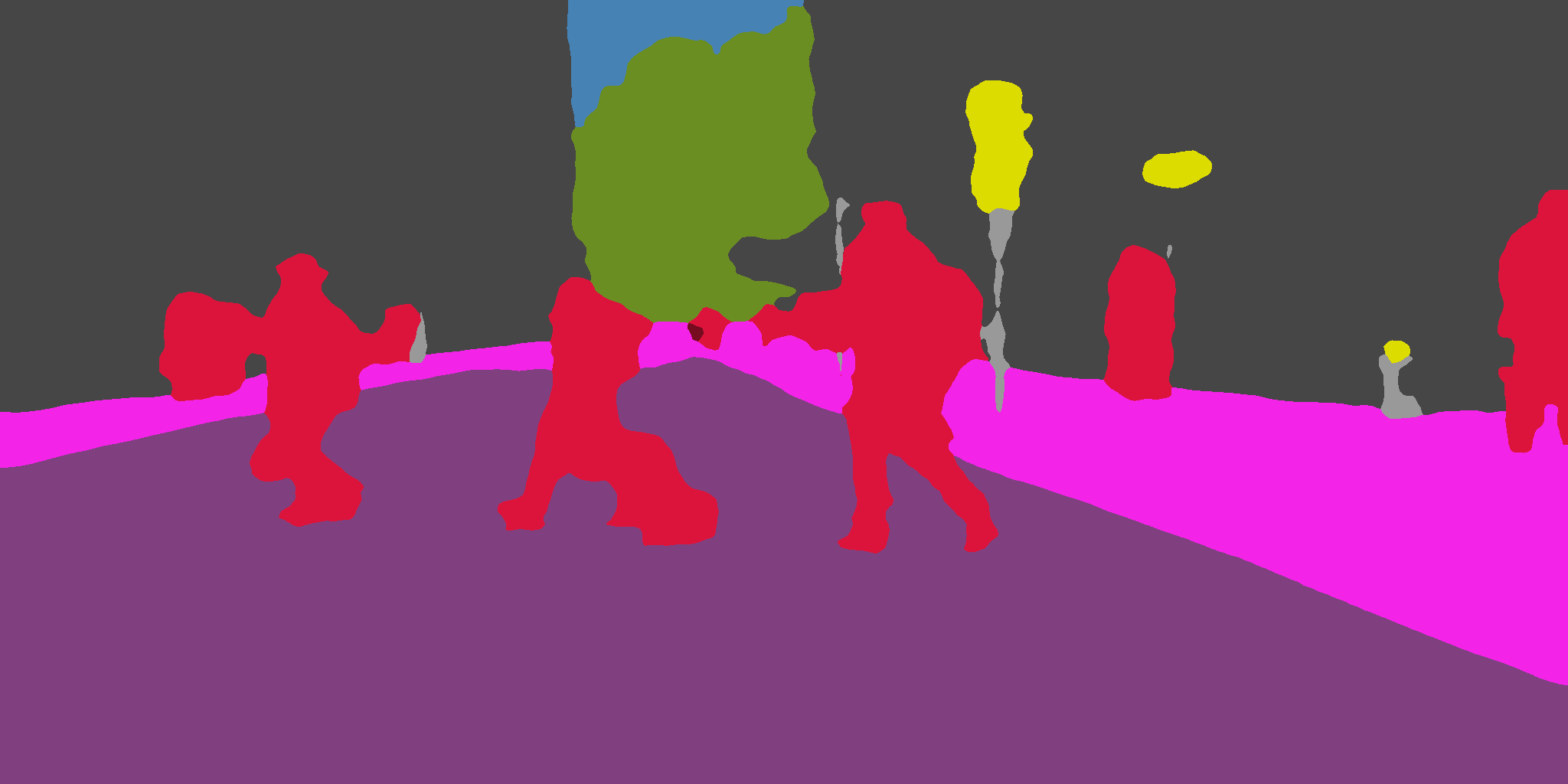}
(d) SANet 
\end{minipage}	
\caption{Scene parsing examples on Cityscapes validation set.}
\label{FIG:CITYSCAPES}
\end{figure*}

For this dataset, the ground-truth of test images are withheld by the organizers, so all methods can only be tested by submitting the results to the evaluation server. The overall comparisons of our model with some recently proposed methods are summarized in Table \ref{TB:CITYSCAPES}. Among the test results on the evaluation server, our proposed SANet outperforms all previous methods in terms of the class IoU. The per-class scene parsing results are reported in Table \ref{TB:PER-CITYSCAPES}. From the table, we can see that even without the pre-training on the 20,000 coarse-labelled images, our method still achieves very promising results. We also illustrate some example results for the scene parsing visualization in Figure \ref{FIG:CITYSCAPES}. Since our model can automatically leverage the spatial and semantic information from multi-layer feature maps, some small objects (such as traffic signs) are accurately segmented by SANet, which demonstrates that the proposed feature aggregation module can well deal with the spatial-aware learning tasks.  

\begin{table*}[t]
\centering \small
\caption{Per-class results on Cityscapes testing set.}
\label{TB:PER-CITYSCAPES}
\setlength{\tabcolsep}{4pt}
\renewcommand{\arraystretch}{0.5}
\begin{tabular}{|c|ccccccccccccccccccc|}
\hline
{\bf Method}	&\rot{road} &\rot{sidewalk} &\rot{building} &\rot{wall} &\rot{fence} &\rot{pole} &\rot{traffic light} &\rot{traffic sign} &\rot{vegetation} &\rot{terrain}  &\rot{sky} &\rot{person} &\rot{rider} &\rot{car} &\rot{truck} &\rot{bus} &\rot{train} &\rot{motorcycle} &\rot{bicycle}\\
\hline
SegNet\cite{TPAMI:SEGNET} 	&96.4 &73.2 &84.0 &28.5 &29.0 &35.7 &39.8 &45.2 &87.0 &63.8 &91.8 &62.8 &42.8 &89.3 &38.1 &43.1 &44.1 &35.8 &51.9 \\
LDN-121\cite{ICCV17:LADDER} &97.4 &80.2 &92.0 &47.6 &53.9 &64.6 &72.8 &76.3 &92.8 &66.4 &95.5 &83.8 &66.1 &94.3 &55.6 &70.3 &67.0 &62.1 &73.0\\
ResNet-38\cite{PR:WD_RESNET} &98.5 &85.7 &93.1 &55.5 &59.1 & 67.1 &74.8 &78.7 &93.7 &72.6 &95.5 &86.6 &69.2 &95.7 &64.5 &78.8 &74.1 &69.0 &76.7   \\
SAC\cite{ICCV17:SAC} 	&98.7 &86.5 &93.1 &56.3 &59.5 &65.1 &73.0 &78.2 &93.5 &72.6 &95.6 &85.9 &70.8 &95.9 &71.2 &78.6 &66.2 &67.7 &76.0 \\
Alex et al.\cite{CVPR18:MT} &98.4 &85.2 &92.8 &54.1 &60.8 &62.4 &73.4 &77.5 &93.3 &71.5 &95.1 &84.9 &69.5 &95.3 &68.5 &86.2 &80.0 &67.8 &75.6 \\
RefineNet\cite{CVPR17:REFINENET} &98.2 &83.3 &91.3 &47.8 &50.4 &56.1 &66.9 &71.3 &92.3 &70.3 &94.8 &80.9 &63.3 &94.5 &64.6 &76.1 &64.3 &62.2 &70.0  \\
PSPNet\cite{CVPR17:PSPNET} &98.6 &86.6 &93.2 &58.1 &63.0 &64.5 &75.2 &79.2 &93.4 &72.1 &95.1 &86.3 &71.4 &96.0 &73.5 &90.4 &80.3 &69.9 &76.9 \\
AAF \cite{ECCV18:AAF}	&98.5 &85.6 &93.0 &53.8 &59.0 &65.9 &75.0 &78.4 &93.7 &72.4 &95.6 &86.4 &70.5 &95.9 &73.9 &82.7 &76.9 &68.7 &76.4 \\
DANet \cite{CVPR19:DANET} &98.6 &86.1 &93.5 &56.1 &{\bf 63.3} &{\bf 69.7} &{\bf 77.3} &{\bf 81.3} &{\bf 93.9} &72.9 &95.7 &87.3 &{\bf 72.9} &96.2 &{\bf 76.8} &{\bf 89.4} &{\bf 86.5} &{\bf 72.2} &{\bf 78.2} \\
\hline
SANet (ours)	&{\bf 98.7} &{\bf 87.1} &{\bf 93.6} &{\bf 61.6} &62.4 &68.1 &75.9 &79.5 &93.8 &{\bf 73.1} &{\bf 95.8} &{\bf 87.3} &71.5 &{\bf 96.2} &71.9 &88.1 &86.1 &69.4 &77.2 \\
\hline
\end{tabular}
\end{table*}

\begin{table}[t]
\centering 
\caption{Overall results on Cityscapes testing set.}
\label{TB:CITYSCAPES}
\begin{tabular}{|c|cccc|}
\hline
{\bf Method}	&IoU cla. &iIoU cla. &IoU cat. &iIoU cat.  \\
\hline
LDN-121\cite{ICCV17:LADDER} &74.3 &51.6 &89.7 &79.5 \\
SAC\cite{ICCV17:SAC} &78.1 &55.2 &90.6 &78.3 \\
RefineNet\cite{CVPR17:REFINENET}	&73.6 &47.2 &87.9 &70.6   \\
Yu et al. \cite{CVPR18:DLA} &75.9 &- &- &- \\
Alex et al.\cite{CVPR18:MT}	&78.5 &57.4 &89.9 &77.7   \\
AAF \cite{ECCV18:AAF}	&79.1 &56.1 &90.8 &78.5 \\
PSPNet\cite{CVPR17:PSPNET}	&78.4 &56.7 &90.6 &78.6   \\
PSANet\cite{ECCV18:PSANET} &78.6 &- &- &- \\
Pad-Net\cite{CVPR18:PAD_NET} &80.3 &58.8 &90.8 &78.5 \\
\hline
SANet (ours)	&{\bf 80.9} &{\bf 59.6} &{\bf 91.4} &{\bf 80.2}   \\
\hline
\end{tabular}
\end{table}

\subsection{Discussions}

The goal of the proposed spatial-semantic feature aggregation module is to learn better feature maps for dense classification. Furthermore, due to the use of the proposed feature aggregation module, the deep parsing network is much easier to train because the gradients can be passed to multiple learning blocks, achieving to a faster training process. So the proposed feature aggregation module is an excellent alternative compared to simple skip-connections or auxiliary losses in the deep scene parsing models. The experimental results have proved the effectiveness of our module. In some other spatial-aware learning tasks such as object detection \cite{NIPS15:FASTER_RCNN,CVPR18:RN} and tracking \cite{CSR:TRACK}, the global feature representation prior that leverages the spatial and semantic information usually obtains superior performance, so it is worth trying to apply the memory cell on multiple feature outputs in deep neural networks to improve the accuracy.

\section{Conclusion}\label{SEC:CONCLUSION}

We have presented a novel spatial-semantic feature selection module for supervised scene parsing. The extra module can select the useful components in multiple feature outputs and aggregate them to form a discriminative global feature representation for accurate pixel-label prediction. Integrating the feature selection module into deep parsing networks also forms a strong supervision to effectively suppress the over-fitting problem, making the parsing network easy to train. Extensive experiments with very promising results on four public scene parsing datasets demonstrate the effectiveness of our SANet. We believe the proposed spatial-semantic feature aggregation module can also benefit the related spatial-aware learning techniques in the community.

\bibliographystyle{IEEEtran}
\bibliography{reference}

\end{document}